\documentclass[11pt]{article}

\usepackage[preprint]{acl}
\usepackage{enumitem} 
\usepackage{float}
\usepackage{times}
\usepackage{latexsym}

\usepackage[T1]{fontenc}
\usepackage{tabularx}
\usepackage{pdflscape}
\usepackage{CJKutf8}

\usepackage[utf8]{inputenc}
\usepackage[dvipsnames]{xcolor}
\usepackage{microtype}

\usepackage{inconsolata}
\usepackage{amsmath}
\usepackage{graphicx}
\usepackage[size=tiny]{todonotes}
\usepackage[skip=2pt]{caption}
\usepackage{booktabs, multirow} 
\usepackage{xcolor,colortbl} 
\usepackage{subcaption}
\usepackage{color,soul}
\title{Beyond Reproduction: A Paired-Task Framework for Assessing LLM Comprehension and Creativity in Literary Translation}

\newcommand{\zr}[1]{\textcolor{black}{#1}}

\author{Ran Zhang \\
Natural Language Learning Group (NLLG) \\
School of Business Informatics and Mathematics \\
University of Mannheim \\
\texttt{ran.zhang@uni-mannheim.de} 
 \And  ... \And
         Author n \\ Address line \\ ... \\ Address line}
\author{
  \textbf{Ran Zhang\textsuperscript{1,2,5}},
  \textbf{Steffen Eger\textsuperscript{2,5}},
  \textbf{Arda Tezcan\textsuperscript{3}} \\
  \textbf{Wei Zhao\textsuperscript{4}},
  \textbf{Simone Paolo Ponzetto\textsuperscript{1}},
\textbf{Lieve Macken\textsuperscript{3}}\\
  \textsuperscript{1}
University of Mannheim, Data and Web Science Group \\
\textsuperscript{2}University of Technology Nuremberg (UTN), Department Engineering \\
\textsuperscript{3}University of Gent, Department of Translation, Interpreting and Communication \\
  \textsuperscript{4}University of Aberdeen, Department of Computing Science\\
  \textsuperscript{5}Natural Language Learning and Generation (NLLG) Lab
\\
  \small{
  Correspondence: \href{mailto:ran.zhang@uni-mannheim.de}{ran.zhang@utn.de}} \ \ 
}

\begin{document}
\maketitle
\begin{abstract}
Large language models (LLMs) are increasingly used for creative tasks such as literary translation. Yet translational creativity remains underexplored and is rarely evaluated at scale, while source-text comprehension is typically studied in isolation, despite the fact that, in professional translation, comprehension and creativity are tightly intertwined. We address these gaps with a paired-task framework applied to literary excerpts from 11 books. Task 1 assesses source-text comprehension, and Task 2 evaluates translational creativity through Units of Creative Potential (UCPs), such as metaphors and wordplay. Using a scalable evaluation setup that combines expert human annotations with UCP-based automatic scoring, we benchmark 23 models and four creativity-oriented prompts. Our findings show that strong comprehension does not translate into human-level creativity: models often produce literal or contextually inappropriate renderings, with particularly large gaps for the more distant English--Chinese language pair. Creativity-oriented prompts yield only modest gains, and only one model, Mistral-Large, comes close to human-level creativity (0.167 vs.\ 0.246). Across all model–prompt combinations, only three exceed a creativity score of 0.1, while the rest remain at or near zero.\footnote{Our code and data are available at \url{https://github.com/NL2G/Beyond-Reproduction}}
\end{abstract}

\section{Introduction}
\begin{figure*}
    \centering
    \includegraphics[width=1\linewidth]{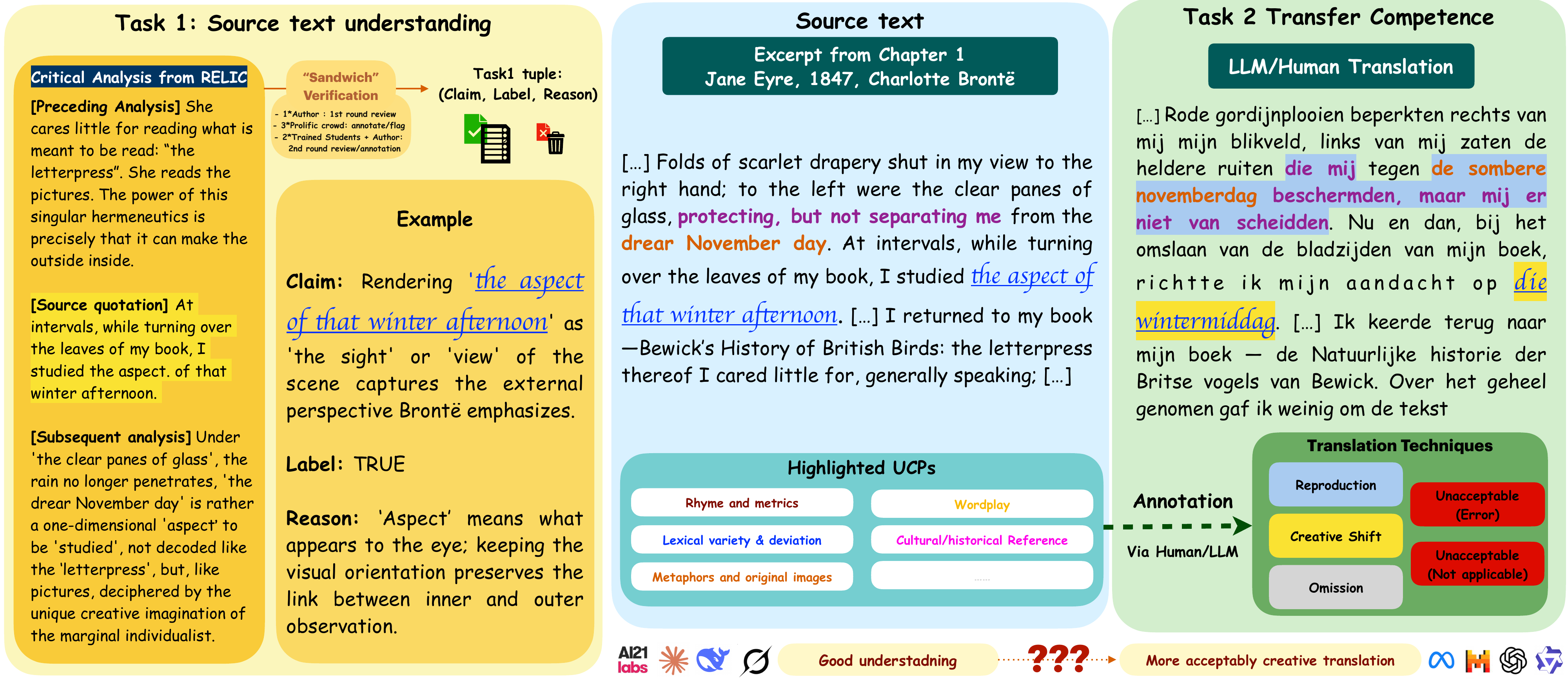}
    \caption{Construction of the paired task linking source understanding (Task 1) and creative transfer (Task 2). Task 1 contains interpretive claims derived from critical analysis under three-stage ``Sandwich'' verification (1 author --> 3 crowdworkers --> 2 trained students + author; numbers indicate annotator counts). Task 2 evaluates translational creativity: text colors indicate Unit-of-Creative-Potential (UCP) types, and highlight colors indicate annotated translation techniques: \textcolor{blue}{\hl{\textbf{die wintermiddag}}} is a creative shift for \textcolor{blue}{\textbf{the aspect of that winter afternoon}} (lexical variety). \zr{Task 1 is designed to assess local contextual interpretation relevant to the comprehension and translation of UCPs. The underlined examples highlight the corresponding spans that link Task 1 to Task 2.}}
    \label{fig:overview}
\end{figure*}

Traditionally, creativity has been regarded as an essentially human capacity~\cite{marrone2024does, bender-2021-stochastic-parrots, role_of_creativity, art_artifice}. With the advent of large language models (LLMs) and the broader debate around artificial intelligence systems, however, this assumption is increasingly being challenged~\cite{koivisto2023best, lu2024llm, lin2025creativity}. On the one hand, LLMs have been described as ``stochastic parrots'' relying on large-scale memorization and probabilistic pattern-matching rather than understanding or intentional creativity~\cite{bender-2021-stochastic-parrots}. On the other hand, claims of human-parity performance on creative tasks are becoming more common~\cite{wu2025perhaps,orwig2024language, qiu-hu-2025-deep}.  At the same time, machine translation (MT), especially when powered by LLMs, is now frequently used to produce or assist translations, including literary ones, a development with substantial implications for readers~\cite{gerrits-arenas-2025-mt,guerberof2020impact}, translators~\cite{macken-2024-machine,macken-etal-2025-role}, and the translated literature itself~\cite{budimir-2025-challenge}. 

Translation scholars view the prominent role of creativity in literary translation as foundational, rather than a mere supplement~\cite{boase2016practices, role_of_creativity}. \citet{guerberof2020impact, arenas-toral-2022-creamt, guerberof-arenas-2024-increc} identify certain ``units of creative potentials'' (UCPs) in the source text as places where translators must move beyond routine solutions. Creativity in this context is the ability to depart from the literal rendering of the source texts and to create a context-appropriate translation~\cite{bayer2009translational, guerberof2022creativity}. Yet, current research on literary translation evaluation largely overlooks these aspects: (1) Most recent studies focus on appropriateness and accuracy, leaving translational creativity almost entirely out of scope~\cite{zhang-etal-2025-good,zhang-etal-2025-litransproqa}, a particularly concerning limitation in light of claims that these systems have reached human parity~\cite{wu2025perhaps}; (2) LLM creativity itself remains under-specified and under-evaluated in NLP: existing work on translational creativity is small-scale yet costly, restricted to comparisons of one or two traditional MT systems, and challenging to scale up~\cite{gerrits-arenas-2025-mt,guerberof2022creativity,macken-etal-2025-role}; 
(3) by focusing solely on the translation \emph{product} or by studying comprehension in isolation~\cite{karpinska-etal-2024-one,sui-etal-2025-kristeva}, most evaluations ignore the underlying \emph{process} dimension. It remains unclear to what extent models understand literary texts and whether that understanding enables them to make context-sensitive creative choices.  
Although LLMs can generate large volumes of fluent, inexpensive translations, we still know relatively little about how they handle the creative demands specific to literary texts.

In this paper, we address these gaps through three main contributions.
(1) To examine the translation process in LLMs, we introduce \textbf{two complementary tasks} (shown in Figure~\ref{fig:overview}): one that targets source-text understanding surrounding UCPs and another that focuses on producing nuanced, context-sensitive translations containing UCPs. We work with literary excerpts from 11 books spanning diverse genres and styles, focusing on two distinct language pairs: English–Chinese and English–Dutch. 
(2) We introduce \textbf{a scalable, expertise-driven evaluation framework} that jointly assesses both appropriateness and creativity. It is supported by over 1000€ worth of human evaluation and can be readily extended to additional models and translation configurations (e.g., alternative prompt designs).
(3) We provide \textbf{a systematic comparison} of 23 models of various sizes across two tasks, covering multiple architectures and model families, and exploring creativity-oriented prompt variants. In total, we analyze over 1k translations and more than 7{.}5k automatically evaluated UCPs to study translational creativity.

Our results show that strong comprehension does not automatically yield human-level creativity, even though the two are moderately correlated. For comprehension (Task~1), scaling up models or adding explicit ``thinking'' modes provides only limited benefits. For creative transfer (Task~2), LLMs often produce literal or contextually inappropriate translations, showing a clear gap from human creativity, especially for the more distant English--Chinese pair. Automatic evaluation reinforces this pattern: creativity-oriented prompting shifts behavior only mildly, and most models remain in the low-creativity region, with only one system (Mistral-Large) closer to human creativity (0.167 vs.\ 0.246). Across all models and prompts, only three systems exceed a creativity score of 0.1; the rest peak between roughly $-0.10$ and $0.03$. Overall, these findings show that (1) current LLMs fall well short of human creative performance, and (2) effective creative transfer requires both strong contextual understanding and careful prompting, conditions under which many models still struggle to produce contextually grounded creative output.

\section{Related work}
\paragraph{Literature comprehension and reasoning.}
Literary works such as novels are creative productions that extend beyond conventional boundaries of cultural history and language~\cite{boase2016practices}. Their length, complex plots, and intertwined character relationships make them a challenging testbed for evaluating language models' comprehension and reasoning~\cite{hamilton2025too, ahuja2025finding}. $\infty$-Bench \cite{zhang-etal-2024-bench} equip classic novels with multiple-choice questions, open-ended question answering (QA), and summarization tasks to evaluate model performance on contexts exceeding 100k tokens. NovelQA~\cite{wang2025novelqa} similarly combines multiple-choice and open-ended question sets, using LLM-based evaluators for open-ended responses. NovelHopQA~\cite{gupta-etal-2025-novelhopqa} targets multi-hop reasoning in long narratives, varying context length, and hop depth. Beyond long-context comprehension, KRISTEVA~\cite{sui-etal-2025-kristeva} introduces close reading as a benchmark for interpretive reasoning through progressively challenging multiple-choice tasks. \citet{karpinska-etal-2024-one} propose NoCha with 1,001 true/false claim pairs mixing human- and model-generated statements. NoCha covers 67 recently published English books, with 48\% of claims requiring global reasoning over the entire novel. 

Unlike prior benchmarks that treat literary tasks as standalone comprehension challenges, our work targets the gap between comprehension and transfer/generation capacity. We propose a dataset of claim–reasoning pairs derived from English literary excerpts. This dataset is further enriched with Chinese and Dutch translations in which trained annotators have manually labeled instances of translation techniques, such as creative shifts and reproductions. Crucially, we position source comprehension as a prerequisite for literary translation. We thus investigate how comprehension affects creative capacity when translating inherently creative source texts.

\paragraph{LLMs for creative tasks and creativity evaluation.}
The increasing capabilities of LLMs have motivated extensive research on their application to creative tasks, including poetry generation~\cite{zhang2024llm, belouadi2023bygpt5}, literary translation~\cite{zhang-etal-2025-good, wu2025perhaps}, and creative writing~\cite{hou2025creativityprism, marco-etal-2025-small}. To enhance LLM performance in creativity-intensive settings, prior work has explored a diverse set of techniques, including adversarial fine-tuning~\cite{zhang2024llm}, decoding adjustments such as temperature-based sampling~\cite{peeperkorn2024temperature}, prompt engineering~\cite{zhao2025assessing, zhang-etal-2025-litransproqa}, and multi-agent systems~\cite{lin2025creativity, wu2025perhaps}. 

Despite substantial exploration in creative generation and quality evaluation~\cite{zhang-etal-2025-good, zhang-etal-2025-litransproqa}, creativity evaluation remains underemphasized and theoretically underdeveloped. Some studies operationalize creativity largely as novelty~\cite{lu2025ai, zhang2025noveltybench}, overlooking its multidimensional nature. Other work adapts psychometric measures from psychology, such as the Remote Associates Test, reapplied to LLMs to generate parallel association chains~\cite{qiu-hu-2025-deep}, and modified Torrance Tests of Creative Thinking~\cite{zhao2025assessing, art_artifice, lu2024llm} to assess fluency, flexibility, originality, and elaboration. While these psychometric approaches offer task-agnostic and easily deployable evaluation tools, several studies report contradictory or unstable findings across domains~\cite{10.1145/3167476, art_artifice, hou2025creativityprism}, raising concerns about their generalizability.

To address these limitations, our work adopts a task-specific approach to creativity evaluation grounded in psychological theory. We adopt the definition of creativity as a combination of novelty and usefulness~\cite{bayer2009translational} and utilize the taxonomy proposed by the CREAMT project~\cite{arenas-toral-2022-creamt}, which has been widely applied in literary translation research~\cite{guerberof2020impact, guerberof2022creativity, Guerberof-Arenas_Valdez_Dorst_2024, gerrits-arenas-2025-mt}. This framework enables a more precise and context-sensitive assessment of creative translation choices, addressing the gaps left by task-agnostic creativity metrics. 

Our work differs substantially from CREAMT and other existing works by linking translational creativity to a preceding comprehension task using a diverse set of books spanning multiple genres and styles. More importantly, our framework is designed for scalability, allowing us to move beyond previous explorations of translational creativity (which were limited to traditional MT): we compare 23 LLMs under multiple prompting conditions, a setting that, to the best of our knowledge, has not been extensively explored.
\section{A paired-task framework to evaluate LLM literary translation}
We propose a framework for assessing LLM literary translation competence through two complementary tasks: (1) understanding the literary source text, and (2) recreating it in the target language with comparable interpretive and aesthetic value. The full workflow is depicted in Figure~\ref{fig:overview}, with key steps outlined below. The first task evaluates interpretive reasoning, such as cultural background, author style, and linguistic difficulties, while the second examines how translation techniques are applied to solve the actual translation problems and their connection to creativity.

\subsection{Task 1: source text understanding}
\paragraph{Dataset construction.}
To construct datasets for the source-text understanding task, we use a pipeline that combines literary critical analysis, automatic claim generation, expert review, and crowd-based validation. This process ensures that each test question contains a stable interpretive judgment grounded in the text and remains scalable across works. We use the RELIC dataset~\cite{thai-etal-2022-relic} as our primary source for generating interpretive claims. Each RELIC entry contains (i) quotation from a literary work and (ii) critical analysis surrounding that quote: [\textit{preceding analysis},\ \textit{quote},\ \textit{subsequent analysis}]. The quotation appears in their original sequence and can be mapped to longer excerpts (e.g., Chapter 1).
While our study focuses on English source texts, the methodology can be extended to other languages with comparable critical corpora. 

\begin{itemize}[leftmargin=0.8em, topsep=-0.01pt, itemsep=-1pt, parsep=-1pt, partopsep=-0.01pt]
\setlength{\itemsep}{-0.05pt}
\setlength{\parskip}{-0.05pt}
    \item \textbf{Step 1 -- Literary excerpt selection}: 
    We select excerpts (paragraphs) from early chapters, especially Chapter 1, as opening paragraphs introduce key characters, tone, and setting—providing a natural starting point for evaluating textual understanding. To identify segments with interpretive or translational challenges, we use automated analysis with multiple LLMs (Gemini2.5-Flash, GPT-4o-mini, and	Claude3-Haiku) to flag sentences or paragraphs containing features associated with translation difficulty defined in \S\ref{apdx:translation_diff} (appendix). Because translation difficulty is language-dependent, automated detection is intentionally broad. We do not require high accuracy at this stage, since excerpts undergo two rounds of verification: (1) by the first author during selection and (2) by two trained students with backgrounds in linguistics or literary translation during Task 2 annotation.
    \item \textbf{Step 2 -- Claim construction}: 
    Building on findings that LLMs can generate structured interpretive hypotheses such as arguments \cite{chen-etal-2024-exploring-potential}, we prompt GPT-5 to generate candidate true/false claims based on the selected RELIC entries in Step 1. Each claim reformulates the critic's argument into a concise assertion and includes a justification from either the critical analysis or textual evidence. Task units consist of: (1) the source excerpt, (2) the claim, (3) the true/false label, and (4) a rationale. The first author reviews these claims to ensure they require interpretive inference rather than simple paraphrasing or factual recall, excluding claims based on personal preference or that are trivial.
    \item \textbf{Step 3 -- Crowd-based validation}: three Prolific annotators with self-reported advanced or native English proficiency and regular literature reading experience work on the filtered claims. Each annotator reads the excerpt and claim (with additional context as needed) and classifies it as True, False, or Problematic, with a brief justification. Annotators may use external knowledge, including online historical or cultural information (see guideline in \S\ref{apdx:task1_guideline}). A claim is retained only if a clear majority agrees on True or False with consistent reasoning. Claims flagged as Problematic or differing from the Step 2 label are revised or discarded, ensuring comprehensibility and interpretive validity.
    \item \textbf{Step 4 -- Expert re-validation}: Claims flagged in Step 3 as Problematic or showing disagreement with the initial generated label are reviewed by the first author and two annotators trained in linguistics/literary translation. Reviewers determine whether issues stem from subjective interpretability or unclear formulation. Claims based on subjective preference are discarded; conceptually valid but imprecise claims are revised and retested. This validation ensures Task~1 maintains high data quality and interpretive reliability. Annotator statistics are in \S\ref{annotator}.\footnote{We compute agreement by comparing the labels assigned to each item across annotator pairs for 30 instances, fully annotated by the author and by trained student annotators. On average, Prolific annotators show an agreement of 0.65, whereas the first author and the two experienced annotators show an agreement of 0.81.} 
\end{itemize}

\paragraph{Dataset statistics.}
The dataset for Task~1 comprises 299 claims from 11 English classic novels spanning the early nineteenth to mid-twentieth century, including works by Herman Melville, Charles Dickens, Virginia Woolf, Emily Brontë, Charlotte Brontë, Mary Shelley, George Orwell, Oscar Wilde, James Joyce, Margaret Mitchell, and Harriet Beecher Stowe. The full book list is shown in Table \ref{tab:apdx_book} (appendix). These works represent diverse genres: realist and historical novels, Gothic fiction, political allegory, and social protest literature. Of the 299 claims, 140 are labeled true and 159 false. Each claim is anchored in an excerpt averaging approximately 186 tokens. \zr{Based on feedback and agreement among crowdsourced annotators and trained students, we categorize Task 1 difficulty into three levels: Low (full agreement on the label), Medium (at least one differing label), and High (at least two differing labels). In total, 299 instances were annotated, of which 140 (46.8\%) are classified as Low difficulty, 106 (35.5\%) as Medium, and 53 (17.7\%) as High. }

\subsection{Task 2: Transfer competence}\label{sec:stage}
\paragraph{Dataset construction.}
Task~2 evaluates a model's ability to handle literary translation challenges through creative strategies. We adopt the CREAMT taxonomy \cite{arenas-toral-2022-creamt}, grounded in the widely accepted definition of translational creativity as the combination of novelty and usefulness \cite{bayer2009translational}. We use Units of Creative Potential (UCPs) and Creative Shifts (CSs) as analytical tools \cite{guerberof2020impact, guerberof2022creativity, Guerberof-Arenas_Valdez_Dorst_2024, gerrits-arenas-2025-mt}. UCPs are text segments that either contain inherently creative elements or constitute ``problematic units that are deemed to require high problem-solving capacity'' \cite{gopferich2011exploring}. We focus on UCPs such as metaphors and similes, wordplay, idioms, cultural/historical references, rhyme/meter, and lexical deviation. Translators are more likely to apply creative techniques to convey the intended interpretive meaning, aesthetic value, or stylistic effect. Even when translators choose a literal rendering, UCPs still mark points where non-literal options are plausibly needed, thus highlighting potential sites of creativity \cite{bayer2011creative}. The translations of each UCP are annotated for acceptability, creativity, and semantic or imagistic deviation (low, medium, or high) to make labels more interpretable. 

As large-scale annotated resources are lacking and human evaluation remains resource-intensive, Task 2 adopts a two-stage design that balances evaluation quality, scalability, and automation.

\paragraph{Stage 1 -- Dataset construction and human evaluation.}
In the first phase, we construct a small, high-quality annotated dataset for two purposes: (i) as a gold standard for comparing human and LLM translations, and (ii) as a development set for scalable evaluation. Two students trained in literary translation and comparative linguistics manually annotate selected UCPs.\footnote{Following Task~1 (Step~1), we focus on excerpts rich in linguistic effects. UCPs in the English--Chinese setting are finalized through joint discussion between the first author and a Chinese-speaking annotator with expertise in literary translation. Problematic units include those from the LitEval-Corpus~\cite{zhang-etal-2025-good}. To ensure cross-language comparability, we annotate the same UCPs for English--Dutch. In a few cases, these are not strictly UCPs in Dutch context due to closer cultural proximity, but they still require careful translational choices.} Annotators focus on highlighted source segments and their corresponding translations from humans and from a diverse set of LLMs (see Task~2 statistics in \S\ref{sec:task2} and details in \S\ref{apdx:task2_detail}). For each selected UCP, they first judge whether the translation is acceptable:
\begin{itemize}[leftmargin=0.9em, topsep=-0.01pt, itemsep=-1pt, parsep=-1pt, partopsep=-0.01pt]
\setlength{\itemsep}{-0.05pt}
\setlength{\parskip}{-0.05pt}
    \item \textbf{Not acceptable (Error)}: It contains grammatical errors or is overly literal.
    \item \textbf{Not acceptable (Not applicable)}: It is not a valid solution as it distorts the meaning beyond recognition or is ``too creative'' for the target audience, even if grammatically correct.
\end{itemize}

\noindent If the translation is acceptable, annotators label one of the following techniques:
\begin{itemize}[leftmargin=0.9em, topsep=-0.01pt, itemsep=-1pt, parsep=-1pt, partopsep=-0.01pt]
\setlength{\itemsep}{-0.05pt}
\setlength{\parskip}{-0.05pt}
    \item \textbf{Reproduction}: The translation conveys the same idea or image as the highlighted source.
    \item \textbf{Omission}: The highlighted content is absent from the target text.
    \item \textbf{Creative Shift (CS)}: Translations that deviate from a literal rendering of the source text to create a more impactful, natural, or culturally appropriate translation in the target language in a meaningful or stylistically motivated way. 
\end{itemize} 

\paragraph{Stage 2 -- Automatic labeling with LLM-as-a-Judge.}
To scale beyond manual evaluation, the second stage adopts an LLM-as-a-Judge setup. The model receives the same guidelines as human annotators and is prompted to assign a translation-technique label and a brief justification for each instance. Its outputs are then compared against the expert-labeled gold standard to assess reliability. We use Qwen3-Next-80B, a Mixture-of-Experts model designed for complex reasoning, which achieves an F1 score of 0.710 on translation-technique labeling on our En–Zh validation set, using 12.8\% of the human-annotated translation paragraphs (See \S\ref{sec:automatic_eval} for the prompt and evaluation setup).\footnote{For per-label precision, recall, and F1 performance, see Tables \ref{tab:validation} in \S\ref{sec:automatic_eval}.} Because translation evaluation is challenging, the evaluation prompt includes concrete in-context examples from human annotations of the \emph{same} source paragraph and language pair. We prioritize adding more translations from diverse models rather than increasing the number of source excerpts or UCPs. This approach combines expert interpretive insight with automated scalability and supports cross-lingual extension of the benchmark.

\paragraph{Dataset statistics.}\label{sec:task2}
\begin{table}[t]\centering
\resizebox{0.48\textwidth}{!}{ 
\begin{tabular}{lccccccc}\toprule
\multirow{2}{*}{Pair} &\multirow{2}{*}{\#book} &\multicolumn{2}{c}{Source} &\multicolumn{2}{c}{Translation} & \\\cmidrule{3-6}
& &\#src &avg.token &\#tgt &avg.token &\#UCPs \\\midrule
En-Zh &\multirow{2}{*}{8} &\multirow{2}{*}{20} &\multirow{2}{*}{217.8} &78 &378.0 &398 \\
En-Nl & & & &76 &219.3 &374 \\
\bottomrule
\end{tabular}}
\caption{Statistics for Task~2 annotation. \#src and \#tgt denote the number of selected source excerpts and their translations, respectively. \textit{avg.\ token} reports the average token counts for source and target texts, and \textit{\#UCPs} denotes the units of creative potential annotated in total.}
\label{tab:stat_task2}
\end{table}
The dataset for Task~2 includes annotation of UCPs in literary translation for two language pairs: English–Chinese (En–Zh) and English–Dutch (En–Nl); summary statistics are shown in Table~\ref{tab:stat_task2}. For each pair, we select 20 source excerpts from 8 novels due to cost constraints, targeting passages rich in figurative language. The English excerpts average 218 tokens, while the corresponding translations average 378 tokens in Chinese and 219 in Dutch, reflecting structural and stylistic differences between the languages. Each excerpt includes three to four translations (at least one published human translation plus several LLM outputs), resulting in 78 target texts for En–Zh and 76 for En–Nl. In total, we annotate 398 UCPs for En–Zh and 374 for En–Nl,\footnote{For some En–Zh pairs, multiple human translation versions are available, which leads to additional annotated UCPs.} providing a detailed benchmark for how human translators and LLMs handle creative translation challenges. This dataset also serves as a testbed for scalability experiments. The LLM outputs cover a diverse set of contemporary models of varying sizes and providers, including Claude3.7-Sonnet-Thinking, GPT\mbox{-}4o, Gemini2.5-Flash, Grok3-Mini, LLaMA3.3-70B, and Qwen2.5-7B.

We use a minimal baseline prompt for collecting translations at this stage: \textit{``Please translate the following literary excerpt into Chinese/Dutch. Excerpt: [excerpt]''}. This setup approximates typical translation use and allows us to measure models’ default tendency toward literal versus creative solutions, without explicitly steering them toward creativity. In other words, it provides a conservative lower bound on the models' intrinsic translational creativity. We explicitly investigate the effect of creativity-oriented prompting in \S\ref{sec:exp}, where we show how models respond when creative translation is prompted at various levels.

\subsection{Human annotators}\label{annotator}
For Task~1 verification, we employed two expert annotators (approximately 10 hours each at 12€/hour; total 240€), both with backgrounds in linguistics or translation studies. One is a near-native speaker of English and the other a college student majoring in English. In addition, we recruited 33 annotators via Prolific, compensated at 10€/hour, for a total of 195€.

For Task~2, we employed two experienced annotators: one native speaker of Chinese (30 hours at 12€/hour) with over 60 hours of prior experience in literary translation annotation, and one Dutch speaker with a background in literary translation (15 hours at 16€/hour) and about 20 hours of prior annotation experience (combined total 600€). Overall, human annotation for both tasks amounts to approximately 1{,}000€.

\section{Evaluation}\label{sec:exp}
We now present our evaluation methodology, describing how we use the dataset and outlining the experimental setup for our use case.

\subsection{Benchmark methodology}
\paragraph{Task 1.} 
Task 1 uses a set of claim–reasoning pairs to benchmark a diverse set of LLMs. For each model, we collect outputs as a predicted label plus a brief justification. To account for class imbalance, we report macro F1 scores as model performance.

\paragraph{Task 2.}
Evaluation in Task~2 focuses on translation competence at the level of UCPs. For each translation and its annotated UCPs, we assess (i) the joint distribution of acceptability and creativity (via a confusion matrix), and (ii) the distribution of translation techniques, mapped to creativity scores: $S_{\text{creativity}}
= \frac{\#CS 
- \#UN}{\#\text{UCPs}}
$,
where $\#\mathit{CS}$ is the number of creative shifts, $\#\mathit{UN}$ denotes the number of unacceptable solutions (i.e., creativity out of context and error cases) and $\#\mathit{UCPs}$ is the total number of UCPs. Intuitively, $S_{\text{creativity}}$ increases with acceptable creative shifts and decreases with unacceptable solutions. The score lies in $[-1,1]$: $1$ means all UCPs are creative shifts, $-1$ means all are unacceptable, and values below $0$ indicate that unacceptable solutions outnumber creative shifts.

\subsection{Experimental setup}\label{sec:exp_design}
In this work, we use our dataset to benchmark a diverse set of contemporary LLMs and their variants, covering different sizes, architectures, and reasoning modes. Concretely, our pool includes 23 models from eight major families (detail in \S\ref{sec:config}): Jamba, Claude, DeepSeek, Gemini, LLaMA, Mistral, GPT, Qwen, and Grok, and human translations. Model sizes range from a few billion parameters to trillion-scale systems. Architecturally, the set covers state-space models (Jamba variants), dense Transformer models (e.g., Claude3.7-Sonnet, LLaMA, GPT\mbox{-}4o), and Mixture-of-Experts (MoE) models (e.g., DeepSeek, Gemini, Mistral, Qwen3, Grok). For several families, we include paired variants with and without explicit reasoning or “thinking” modes (e.g., Claude3.7\mbox{-}Sonnet and Qwen3), enabling controlled comparisons of reasoning style at a similar scale.

Beyond a minimal base prompt, we design three creativity-oriented prompts that gradually increase the creative freedom: (1) p1 frames the model as a literary translator whose task is to preserve meaning, tone, stylistic effects, imagery, voice, and cultural nuance without explicitly encouraging creative deviation; (2) p2 keeps these requirements but explicitly allows selective creative adjustments when literal renderings would be awkward or lose nuance; and (3) p3 foregrounds the highly creative nature of the source text and explicitly introduces Creative Shifts as a permissible translation technique (see \S\ref{apdx:task2_detail} for prompts and configurations.)

\zr{Additionally, as our benchmark is built from classic literary texts, some models may inevitably have prior exposure to the source material.\footnote{The reviewers raised potential concerns about data contamination. However, in our setting, both the claims and the labels are constructed through substantial human effort by trained students and Prolific annotators, rather than being directly available from the source material. Accordingly, there is no direct leakage of labels or other evaluation targets: knowledge of the source material alone does not reveal the target labels and therefore does not invalidate the evaluation. This distinction is important because prior exposure to source data is inherent to most historical datasets and common across the literature. That said, we are also interested in the effect of source-data exposure itself, which we examine through an adversarial experiment.} To assess the effect of such source exposure on Task 1, we conduct an adversarial ablation study that perturbs key details in the source paragraphs. The results support the overall validity of Task 1 as a test of model comprehension, while identifying a small number of explanation-level cases consistent with source-based memorization (see \S\ref{apdx:ablation} for details).}
\section{Results}
\subsection{Task 1: Source text understanding}

\begin{figure}[t]
    \centering
    \includegraphics[width=.95\linewidth]{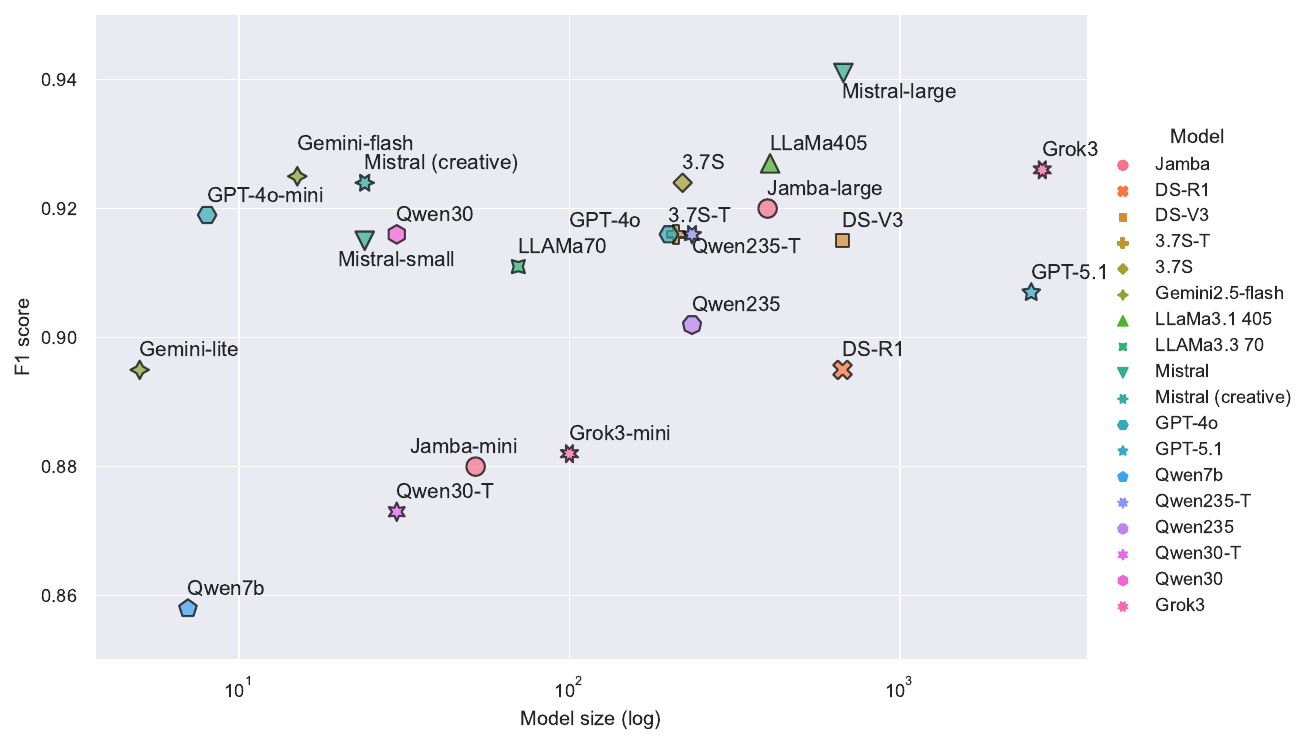}
    \includegraphics[width=.95\linewidth]{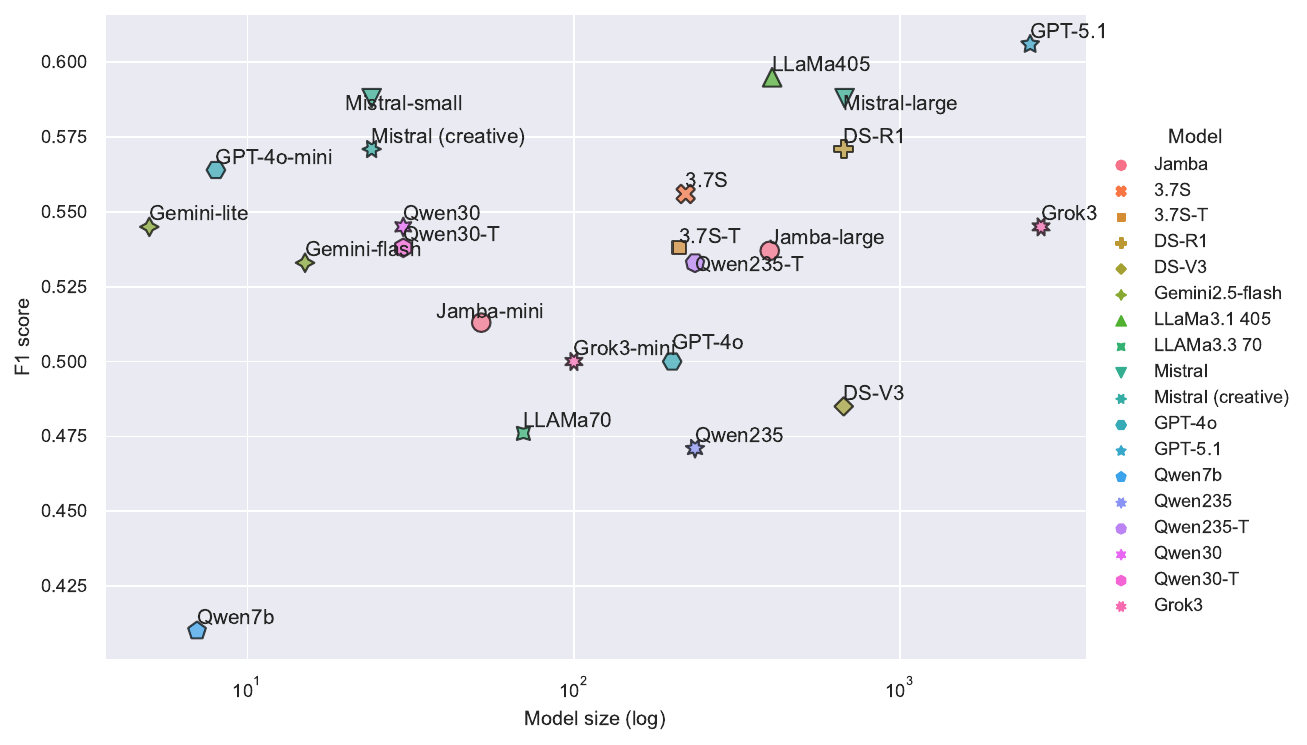}
    \caption{Task~1 performance (F1 score) as a function of model size (log scale). For some proprietary models, parameter counts are approximate and derived from external estimates. \zr{Top: results on all instances; bottom: results on the high-difficulty subset.}}
    \label{fig:task1_size}
\end{figure}
Figure~\ref{fig:task1_size} shows Task~1 performance, measured by macro F1 score versus model size on a log scale, \zr{for all instances (top) and for the high-difficulty subset only (bottom).} For some proprietary models, parameter counts are approximate and based on external estimates.

\paragraph{Scaling laws: bigger, but not always better.}
Across models, F1 scores range from 0.85 to 0.94, \zr{while on the high-difficulty subset they drop to about 0.42-0.60 and exhibit a substantially larger gap in performance.} Model size shows only a mild and non-significant correlation with F1 ($\rho = 0.311$, $p = 0.149$), and many systems deviate from a simple ``bigger is better'' pattern. Within model families, scaling trends mostly hold \zr{in both panels}: ``mini'' variants lag behind their larger counterparts (e.g., Jamba, and Grok3), with GPT\mbox{-}4o\mbox{-}mini a notable exception. Across families, however, size is a weak predictor. Large open models such as Mistral\mbox{-}Large-675B and LLaMA3.1-405B achieve top performance \zr{and remain strong on the hard subset}, but several mid- and small-sized systems (e.g., Gemini2.5\mbox{-}Flash\mbox{-}Lite, Mistral\mbox{-}Small\mbox{-}24B, GPT\mbox{-}4o\mbox{-}mini) reach similar good performance. Conversely, some of the largest models (e.g., DeepSeek\mbox{-}R1, GPT\mbox{-}5.1) underperform relative to smaller ones \zr{on the full set, but their relative performance improves on the high-difficulty subset.}

\paragraph{Mixed results for architecture, training objectives, and reasoning.}
Beyond scale, architecture and training objectives have inconsistent effects on Task~1 performance. State-space models such as Jamba\mbox{-}Large\mbox{-}398B underperform relative to dense Transformer models of similar (or even smaller) size \zr{in both plots}. Mixture-of-Experts (MoE) models exhibit mixed behavior: systems like Mistral-Large and Gemini achieve (near-)frontier performance, whereas other large MoE or hybrid models (e.g., DeepSeek\mbox{-}R1 and GPT\mbox{-}5.1) lag behind smaller dense baselines \zr{on the full set, despite being more competitive on high-difficulty items.} Mistral\mbox{-}Small\mbox{-}24B is exceptionally strong given its comparatively small parameter count, and the variant fine-tuned for creative tasks remains highly competitive \zr{across difficulty levels}. Reasoning-enabled variants also yield non-uniform gains: Qwen3\mbox{-}235B-Thinking improves over Qwen3\mbox{-}235B for both cases, whereas Qwen3\mbox{-}30B-Thinking underperforms Qwen3\mbox{-}30B, and Claude3.7\mbox{-}Sonnet\mbox{-}Thinking (3.7S-T) only slightly improves on the non-thinking version (3.7S), indicating that explicit ``thinking'' modes do not deliver universally higher comprehension performance facing this task, \zr{especially when we focus on the hard cases.}

\subsection{Task 2: Transfer competence}
We report Task~2 results in two parts. First, we present human annotation results from trained linguistics and translation students, offering a detailed view of model behavior on a curated subset. Second, we scale the analysis using an automatic evaluator, enabling systematic benchmarking of translation creativity across many models and prompts.

\paragraph{Analysis based on human annotation.}\label{sec:human-annotation-results}
Table~\ref{tab:joint_human_llm} presents the confusion matrices of acceptability and creativity levels for (a) human translations and (b) the top three LLM systems (with base prompt on Task~2), for English--Chinese (En--Zh) and English--Dutch (En--Nl). Each cell indicates the proportion of UCP translations labeled with that acceptability–creativity combination.

\begin{table}[!htp]\centering
\resizebox{0.485\textwidth}{!}{ 
\begin{tabular}{lcccccccccc}\toprule
\multirow{4}{*}{\cellcolor[HTML]{d9d2e9}\textbf{Level of Acceptability}} &\multicolumn{9}{c}{\cellcolor[HTML]{ead1dc}\textbf{Level of creativity}} \\\cmidrule{2-10}
&\multicolumn{9}{c}{\textbf{a. Human Translation}} \\\cmidrule{2-10}
&\multicolumn{4}{c}{\cellcolor[HTML]{ead1dc}\textbf{En-Zh }} &&\multicolumn{4}{c}{\cellcolor[HTML]{ead1dc}\textbf{En-NL}} \\\cmidrule{2-5}\cmidrule{7-10}
&\cellcolor[HTML]{ead1dc}\textbf{low} &\cellcolor[HTML]{ead1dc}\textbf{medium} &\cellcolor[HTML]{ead1dc}\textbf{high} &\cellcolor[HTML]{ead1dc}\textbf{sum} &&\cellcolor[HTML]{ead1dc}\textbf{low} &\cellcolor[HTML]{ead1dc}\textbf{medium} &\cellcolor[HTML]{ead1dc}\textbf{high} &\cellcolor[HTML]{ead1dc}\textbf{sum} \\\midrule
\cellcolor[HTML]{d9d2e9}\textbf{low} &\cellcolor[HTML]{fffcf1}0.04 &\cellcolor[HTML]{fffffd}0.01 &0.00 &\cellcolor[HTML]{fcfefd}0.04 &\textbf{low} &\cellcolor[HTML]{fff4d3}0.11 &\cellcolor[HTML]{fff4d3}0.11 &\cellcolor[HTML]{fffffc}0.01 &\cellcolor[HTML]{d2ede0}0.23 \\
\cellcolor[HTML]{d9d2e9}\textbf{medium} &\cellcolor[HTML]{fff3d1}0.12 &\cellcolor[HTML]{fffbee}0.04 &\cellcolor[HTML]{fffdf5}0.03 &\cellcolor[HTML]{dcf1e7}0.18 &\textbf{medium} &\cellcolor[HTML]{ffe59d}0.24 &\cellcolor[HTML]{ffedba}0.17 &\cellcolor[HTML]{fffbf0}0.04 &\cellcolor[HTML]{a0d9bd}0.45 \\
\cellcolor[HTML]{d9d2e9}\textbf{high} &\cellcolor[HTML]{ffebb3}0.19 &\cellcolor[HTML]{ffd666}0.38 &\cellcolor[HTML]{ffe9aa}0.21 &\cellcolor[HTML]{57bb8a}0.77 &\textbf{high} &\cellcolor[HTML]{fff4d6}0.10 &\cellcolor[HTML]{fff0c6}0.14 &\cellcolor[HTML]{fff7e0}0.08 &\cellcolor[HTML]{bde5d1}0.32 \\
\cellcolor[HTML]{d9d2e9}\textbf{sum} &\cellcolor[HTML]{b9e3ce}0.34 &\cellcolor[HTML]{a6dbc1}0.42 &\cellcolor[HTML]{d0ecdf}0.24 & &\textbf{sum} &\cellcolor[HTML]{a0d9bd}0.45 &\cellcolor[HTML]{a6dbc2}0.42 &\cellcolor[HTML]{e9f6f0}0.12 & \\
\cmidrule{2-10}
&\multicolumn{9}{c}{\textbf{b. Top-3 LLM per creativity score}} \\
&\multicolumn{4}{c}{\cellcolor[HTML]{ead1dc}\textbf{En-Zh }} &&\multicolumn{4}{c}{\cellcolor[HTML]{ead1dc}\textbf{En-NL}} \\\cmidrule{2-5}\cmidrule{7-10}
&\cellcolor[HTML]{ead1dc}\textbf{low} &\cellcolor[HTML]{ead1dc}\textbf{medium} &\cellcolor[HTML]{ead1dc}\textbf{high} &\cellcolor[HTML]{ead1dc}\textbf{sum} & &\cellcolor[HTML]{ead1dc}\textbf{low} &\cellcolor[HTML]{ead1dc}\textbf{medium} &\cellcolor[HTML]{ead1dc}\textbf{high} &\cellcolor[HTML]{ead1dc}\textbf{sum} \\
\cmidrule{2-5}\cmidrule{7-10}
\cellcolor[HTML]{d9d2e9}\textbf{low} &\cellcolor[HTML]{ffefc3}0.15 &0.00 &0.00 &\cellcolor[HTML]{e4f4ec}0.15 &\textbf{low} &\cellcolor[HTML]{fff8e2}0.07 &\cellcolor[HTML]{fff6db}0.09 &\cellcolor[HTML]{fffffc}0.01 &\cellcolor[HTML]{dff2e9}0.17 \\
\cellcolor[HTML]{d9d2e9}\textbf{medium} &\cellcolor[HTML]{ffebb3}0.19 &\cellcolor[HTML]{fffaeb}0.05 &\cellcolor[HTML]{fffefb}0.01 &\cellcolor[HTML]{ceebdd}0.25 &\textbf{medium} &\cellcolor[HTML]{ffe18d}0.28 &\cellcolor[HTML]{fff1c8}0.14 &\cellcolor[HTML]{fffef9}0.02 &\cellcolor[HTML]{a3dabf}0.44 \\
\cellcolor[HTML]{d9d2e9}\textbf{high} &\cellcolor[HTML]{ffd767}0.38 &\cellcolor[HTML]{ffe9ab}0.21 &\cellcolor[HTML]{fffdf7}0.02 &\cellcolor[HTML]{7ecba5}0.60 &\textbf{high} &\cellcolor[HTML]{ffe59d}0.24 &\cellcolor[HTML]{ffefc1}0.15 &0.00 &\cellcolor[HTML]{acdec6}0.40 \\
\cellcolor[HTML]{d9d2e9}\textbf{sum} &\cellcolor[HTML]{65c194}0.71 &\cellcolor[HTML]{cbeadb}0.26 &\cellcolor[HTML]{feffff}0.03 & &\textbf{sum} &\cellcolor[HTML]{7fcca6}0.60 &\cellcolor[HTML]{b0dfc8}0.38 &0.02 & \\
\bottomrule
\end{tabular}}
\caption{Confusion matrices of acceptability and creativity levels for (a) human translations and (b) the top three LLM systems on Task~2, for English--Chinese (En--Zh) and English--Dutch (En--Nl). Cells show the proportion of UCP translations labeled with each acceptability–creativity combination; row and column sums give the corresponding marginals.}\label{tab:joint_human_llm}
\end{table}
Most human translations of UCPs fall into high acceptability with medium or high creativity. In En--Zh, 77\% of UCP translations are rated highly acceptable, with 38\% at medium and 21\% at high creativity. The En--Nl data exhibits a slightly different pattern, with a higher portion of low-acceptability cases, possibly due to the use of older human translation versions. For both language pairs, humans converge on medium and high acceptability with non-trivial creativity: En--Zh contains roughly twice as many high-creativity decisions as En--Nl, suggesting that more distant language pairs (e.g., En--Zh) may require more creative solutions than closer pairs (e.g., En--Nl).

By contrast, top LLMs produce translations in low-creativity but still acceptable regions. In En--Zh, 60\% of UCP translations are highly acceptable, yet 38\% fall into the high-acceptability/low-creativity cell and only 2\% reach high creativity. In En--Nl, 40\% are highly acceptable, but similarly, only 2\% achieve high creativity. Overall, LLMs favor literal, low-creativity solutions, whereas humans more often yield high acceptability with medium or high creativity. \zr{For examples illustrating these patterns, see \S\ref{sec:apdx_example}.}

\begin{figure*}[t]
\centering
\begin{subfigure}{0.48\textwidth}
    \centering
    \includegraphics[width=1\linewidth,height=4.1cm]{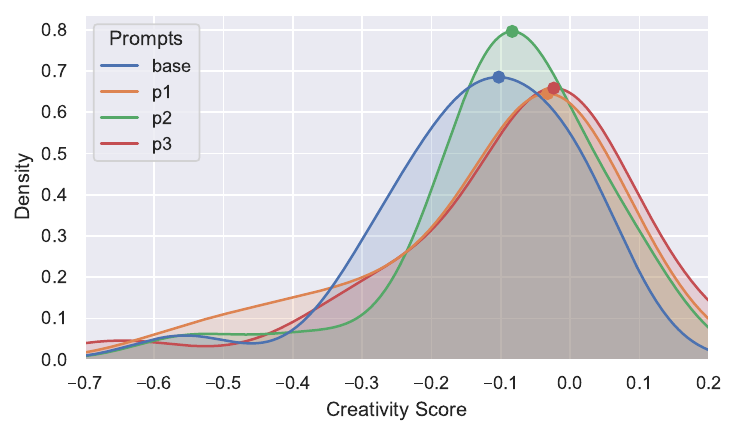}
    \caption{}\label{fig:prompt_eff}
\end{subfigure}\hfill
\begin{subfigure}{0.48\textwidth}
    \centering
    \includegraphics[width=1\linewidth,height=4.1cm]{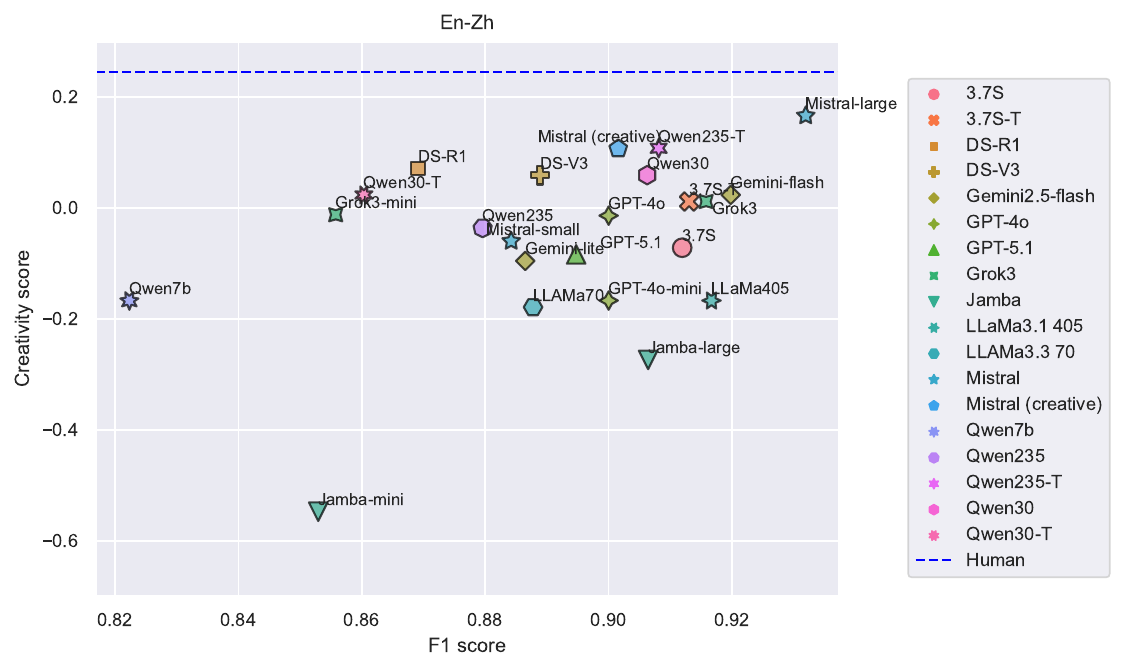}
    \caption{}\label{fig:relation_auto}
\end{subfigure}
\caption{(a) Kernel density estimates of creativity scores for 23 En–Zh models under four prompts (base, P1–P3) obtained via automatic annotation. Prompts encode increasingly explicit creative instructions. ``\scalebox{1}{$\bullet$}'' marks expected values. (b) Source comprehension (Task 1; F1) versus creative transfer (Task 2). The blue line indicates human performance. Each model’s point reflects its best creativity score across the four prompts.}
\end{figure*}

\paragraph{Analysis based on automatic annotation: prompt effects on translational creativity.}
We analyse how prompt framing affects creative translation for 23 models on the En--Zh pair, covering over 7k translated UCPs. Figure~\ref{fig:prompt_eff} shows creativity scores under four prompts (base, P1--P3) varying in how explicitly they encourage creativity. Overall, the score distributions largely overlap (with the base prompt and p2 slightly worse), indicating no clear separation among prompts and no monotonic gains. At the model level, 7 of 23 systems achieve their best score under P3, with only Mistral-Large at 0.167, closer to human creativity at 0.246.\footnote{See Table~\ref{tab:scores} in the appendix for detailed scores from automatic annotation.} Across all models and prompts, only three systems ever exceed 0.1 (Mistral-Large, Mistral-Small–Creative, and Qwen3-235B-Thinking); all others peak between roughly $-0.10$ and $0.03$. Prompting shifts behavior only marginally, and most models remain in the low-acceptability/creativity region. For many systems, greater creative freedom is even counterproductive, leading to ungrounded elaborations that lower scores. DeepSeek-V3.2, Jamba-Large, and LLaMA3.1-405B perform best under the base prompt and degrade under more creative freedom. Overall, prompt framing has modest and uneven effects. Except for Mistral-Large, the rest of the LLMs fall well below human creativity score.

\paragraph{Comprehension–creativity relationship.}
We examine the relationship between comprehension performance (F1) and creative transfer measured by creativity score. Figure~\ref{fig:relation_human} (appendix) shows results based on human evaluation, while Figure~\ref{fig:relation_auto} presents automatic evaluation for the English–Chinese pair (plotting each model's best score across the four prompts). Under human evaluation, comprehension and creativity are positively associated for both language pairs: En–Zh (Spearman $\rho = 0.771$, $p = 0.072$) and En–Nl (Spearman $\rho = 0.714$, $p = 0.047$). Automatic evaluation across 23 models yields a weaker but more significant correlation ($\rho = 0.278$, $p = 0.007$). Overall, better comprehension only mildly supports creative transfer, independent of model size. Human evaluation further highlights language-distance effects: the more distant En–Zh pair exhibits uniformly lower creativity scores and a larger gap from human performance, consistent with evidence that linguistic and cultural distance increases translation difficulty \cite{ploeger2025cross}.

\section{Conclusion}
In this paper, we presented a novel paired-task framework for evaluating literary translation that jointly measures source-text understanding (Task 1) and creative transfer (Task 2) on difficult translation units, combining expert annotation with automatic evaluation to compare models, prompts, and language pairs at scale. Our results show that performance across tasks is only moderately correlated, i.e., \emph{strong comprehension does not reliably lead to creative translations}. Across more than 1k translations and 7.5k UCP evaluations, all but one LLM remain well below human creativity. Scaling and thinking yields only limited gains, and the gap is larger for the more distant English–Chinese pair.

We also provided validity evidence for the evaluation framework. The adversarial ablation study suggests that Task 1 is largely context-grounded, with rare label changes across 200 cases, although a small number of responses still draw on memorized source-text knowledge. For Task 2, the LLM-as-a-judge is more reliable for system-level comparison than for segment-level labeling, supporting its use for model comparison while also warranting caution in fine-grained interpretation.

Overall, these findings indicate that current LLMs have not yet reached human parity in translational creativity, echoing the conclusion of \citet{zhang-etal-2025-good}, who report similar limitations in overall translation quality. At the same time, the modest gains from creativity-oriented prompting suggest that tailored or more advanced prompt engineering remains a promising direction for improving creative generation.

\section*{Limitations}

While we include creativity-oriented prompts, we do not systematically explore the broader prompt design space or advanced methods that could further boost creativity, such as multi-agent systems \cite{lin2025creativity}, decoding exploration, or fine-tuning. Future work could investigate these directions more comprehensively to reveal additional gains in translational creativity.

Although we include both close and distant language pairs, our evaluation is still dominated by high- and medium-resource pairs, and annotated datasets for translational creativity remain scarce overall. This highlights the need to substantially expand the dataset size, cover more languages, and develop dedicated evaluation resources for low-resource languages within the literary domain.

Our corpus primarily consists of well-known classic literary works, which may already be included in LLM pre-training data. This raises potential data-exposure concerns and suggests that our results may represent optimistic upper bounds. Although our ablation study provides evidence that LLM comprehension is driven largely by the given context, we also observe that some models may rely on memorization during reasoning. Extending the framework to newer books, diverse genres, and underrepresented authors, ideally with stricter control over training-data exposure, will be important for future work.

\section*{Ethical Considerations}
We use public-domain classic books and their translations in our dataset, subject to local copyright constraints. For datasets containing copyrighted material, we rely on fair-use principles for research and academic purposes. 

For human evaluation, we obtained informed consent from all participating crowd workers and trained student annotators, and their contributions are recorded anonymously without any protected demographic or personal information.

\paragraph{Potential risks.}
Potential risks of our dataset include reinforcing biases toward high- and medium-resource languages, while effects on underrepresented low-resource languages remain unknown due to limited coverage. These risks call for careful, responsible use and interpretation of benchmark results, as well as future work toward more comprehensive and inclusive dataset coverage.

\paragraph{Licensing and intended use.}
Our implementation builds on components from fastllm-kit 0.1.9 (MIT License). Our use of these artifacts, as well as the LLMs employed in this work, complies with their respective licenses and usage policies. This release is intended solely for research and evaluation in literary translation.

\paragraph{PII in data.}
We rely exclusively on existing, publicly available book sources and translations. We have not conducted independent, systematic checks for personally identifiable information (PII) in these datasets, as we consider this the responsibility of the original writer and publisher. Because our work uses literary excerpts, some texts may contain offensive language; in this context, such content reflects the original material and can, in some cases, be directly relevant to the research questions.

\paragraph{Packages.}
We use the fastllm-kit 0.1.9. The following important dependencies are used: scipy 1.10.1, scikit-learn 1.2.2, seaborn 0.11.2, transformers 4.40.0, and openai 1.56.2 (for API calls).
\section*{Acknowledgments}
We thank the anonymous reviewers for their constructive feedback, which substantially improved this work. We are also grateful to our student annotators for their dedicated contributions, and to the participants recruited via Prolific. The NLLG Lab gratefully acknowledges support from the Federal Ministry of Education and Research (BMBF) via the research grant ``Metrics4NLG'' and the German Research Foundation (DFG) via the Heisenberg Grant EG 375/5-1.

\bibliography{custom}

\appendix

\section{Dataset construction}
\subsection{Task 1 details}
\subsubsection{Translation difficulty}\label{apdx:translation_diff}
Tables \ref{tab:apdx_difficulty1} and \ref{tab:apdx_difficulty2} contain the prompt with the definition of translation difficulties, i.e., Units of Creative Shift (UCPs).  
\begin{table}
    \centering
    \begin{tabularx}{0.49\textwidth}{X}
    \toprule
You are an expert in literary translation. Your task will be to carefully read and analyse the given English literary texts as if you were translating them, and to identify the different (stylistic) features at a local level and determine whether these features pose translation problems.\\
\text{[Translation problem: definition]}\\
1. Metaphors and original images: A metaphor is a figure of speech in which a word or phrase is applied to an object or action to which it is not literally applicable, but helps explain an idea or make a comparison.
E.g., Time is a thief. \\
2. Comparisons/simile: A figure of speech involving an explicit comparison of one thing with another thing of a different kind used to make a description more empathic or vivid (often using words like “like”, “-like”, “as”, and “-looking”. 
E.g. Her smile was as bright as the sun.\\
3. Wordplay: Wordplay refers to the clever and humorous use of words in a way that exploits their multiple meanings, sounds, or structures. E.g., He had a photographic memory but never developed it.
E.g., I scream for ice cream. \\
4. Cultural and historical references: References to culture-specific items or historical events, figures, or practices.
E.g., Achilles heel, French Revolution \\
5. Idiomatic phrases: An idiom is a phrase or expression that largely or exclusively carries a figurative or non-literal meaning.
E.g., piece of cake, bite the bullet, break the ice \\
6. Rhyme and metrics: Use of rhyme and rhythmic patterns
E.g. It was the best of times, it was the worst of times, it was the age of wisdom, it was the age of foolishness, it was the epoch of belief, it was the epoch of incredulity, it was the season of light, it was the season of darkness, it was the spring of hope, it was the winter of despair. (Charles Dickens, A Tale of Two Cities) \\
7. Lexical variety \& lexical deviation: The use of a wide range of vocabulary, including infrequent words, or departure from conventional meanings and uses of words.
E.g., a well-to-do girl of about thirteen (Sarah Waters - Night Watch). E.g., A great sport ... old Evie! (Agatha Christie - The Mysterious Affair at Styles) \\
8. Proper names: Names of people, places, and organizations.
E.g., Albus Dumbledore (Harry Potter - J.K. Rowling), Oompa-Loompas (Roald Dahl - Charlie and the Chocolate Factory). \\
\bottomrule
    \end{tabularx}
    \caption{Translation difficulty (1)}
    \label{tab:apdx_difficulty1}
\end{table}
\begin{table}
    \centering
    \begin{tabularx}{0.48\textwidth}{X}
\toprule
Now read the following text and determine whether it contains any translation problems or not. \\
{text} \\
Please return the answer in the following format: \\
{"problem\_type": "metaphor", "problem\_text": "Time is a thief.", "reason":"explain why it is a problem?"}, {"problem\_type": "simile", "problem\_text": "Her smile was as bright as the sun.", "reason":"explain why it is a problem?"} \\
\bottomrule
    \end{tabularx}
    \caption{Translation difficulty (2)}
    \label{tab:apdx_difficulty2}
\end{table}

\subsubsection{Task 1 annotation guideline}\label{apdx:task1_guideline}
\noindent\textbf{Overview}
You will evaluate whether a claim (a short statement) is True, False, or Problematic, based on the provided source excerpt and the main quote.

\noindent\textbf{For each claim, you must:} \\
Read the source carefully \\
Decide if the claim is True, False, or Problematic. \\
Provide a short reasoning. \\
Assign a confidence score (1–5). \\
\noindent\textbf{Step 1: Read the source carefully.} \\
Read the excerpt thoroughly. Pay special attention to the highlighted quote. This is the key evidence for evaluating the claim.\\
\noindent\textbf{Step 2: Judge the claim}\\
Label each claim with one of the following:
\begin{itemize}
    \item  True: The claim is factually correct and is clearly supported by the source excerpt/quote or corresponding background.
    \item False: The claim contradicts or misrepresents the source excerpt/quote.
    \item Problematic: The claim cannot be reliably judged because of:
Grammatical errors or broken syntax (unclear meaning).
Vagueness/incompleteness (too little detail to evaluate).
Irrelevance (unrelated to the given source/quote).
\end{itemize}
\noindent\textbf{Step 3: Provide reasoning} \\
Explain why you labeled the claim as True, False, or Problematic. Cite or paraphrase the relevant part of the source/quote whenever possible. Keep your explanation short and clear. \\
\noindent\textbf{Step 4: Confidence score}
\\Rate how confident you are in your judgment on a scale of 1–5:
\begin{itemize}
    \item 5 = Very confident → strong evidence, no ambiguity.
    \item 4 = Fairly confident → mostly clear, minor uncertainty.
    \item 3 = Unsure → weak evidence or some ambiguity.
    \item 2 = Low confidence → claim is hard to interpret, limited evidence.
    \item 1 = No confidence  → claim is unable to interpret, no evidence.
\end{itemize}

\subsubsection{Book details}
Table \ref{tab:apdx_book} lists the complete set of books.
\begin{table}[!htp]\centering
\resizebox{0.5\textwidth}{!}{ 
\begin{tabular}{lllll}\toprule
\textbf{Title} &\textbf{Author} &\textbf{Style/Genre} &\textbf{Year} &\textbf{Task 2} \\\midrule
Frankenstein &Mary Shelley &Gothic &1818 &Y \\
Wuthering Heights &Emily Brontë &Gothic &1847 &Y \\
Jane Eyre &Charlotte Brontë &Bildungsroman/Gothic &1847 &Y \\
David Copperfield &Charles Dickens &Bildungsroman &1850 &Y \\
Moby-Dick &Herman Melville &adventure &1851 &Y \\
Uncle Tom’s Cabin &Harriet Beecher Stowe &Anti-slavery &1852 &Y \\
The Picture of Dorian Gray &Oscar Wilde &Philosophical/Gothic &1890 & \\
A Portrait of the Artist as a Young Man &James Joyce &Modernist &1916 & \\
Mrs Dalloway &Virginia Woolf &stream-of-consciousness &1925 &Y \\
Gone with the Wind &Margaret Mitchell &Historical &1936 &Y \\
Animal Farm &George Orwell &Political satire &1945 & \\
\bottomrule
\end{tabular}}
\caption{Book list}\label{tab:apdx_book}
\end{table}

\subsection{Task 2 details}\label{apdx:task2_detail}
\subsubsection{Task 2 annotation guideline}
\noindent\textbf{Overview}
Your task is to evaluate how the highlighted part of the source text has been translated. Focus on the highlighted part and its corresponding translation. Use the surrounding text to check the acceptability and coherence of the translation within the whole. The final classification should be based on the highlighted segment, but always consider the whole excerpt to ensure the translation works in context. \\

\noindent\textbf{Step 1: Read the source and translation.} 
Identify the highlighted part in the source text (ST). Locate its corresponding target translation (TT). Read the surrounding text to check whether the translation is acceptable within the full sentence/paragraph.\\
\noindent\textbf{Step 2: Classify the Translation}\\
\textbf{Detailed definitions: }
\begin{itemize}
    \item  Not acceptable: The translation is either erroneous or not applicable.  
    \begin{itemize} 
    \item Error: The translation contains grammar errors of various types or is overly literal. E.g., literally translating "raining cats and dogs" to its mere semantic equivalent “dropping dogs and cats” rather than "heavy rain." 
    \item Not Applicable: If the translation cannot be evaluated as a valid solution (e.g., distorts meaning beyond recognition but does not contain grammar issues, “too creative”; not acceptable to the target audience at all), mark it as Not Applicable (NA). E.g., Raining cats and dogs → A storm of fur filling the heavens. 
    \end{itemize}
\item Reproduction: The translation conveys the same idea or image as the highlighted source. May still vary in style or acceptability, but the meaning is faithfully preserved.
\begin{itemize}
    \item Retention: the TT keeps the original ST term or expression, e.g., un Rioja → a Rioja (borrowed as-is in English)
    \item Specification: the TT keeps the image of the original ST and it either adds information (e.g., explains the source text's image in the footnote) or it spells out the ST, e.g., if there were an acronym in English where you explained in English what it means: UNESCO (United Nations Educational, Scientific and Cultural Organization)
    \item Direct translation: the TT keeps the same semantical and syntactical structure of the ST. This covers literal translation in its broader sense, that is, it is a correct translation, not what a non-professional would call “literal translation”. E.g., She lives in Paris. → Ella vive en París. (She lives in Paris.)
    \item  Official translation: the TT uses a different translation, but this is an existing translation. E.g., the name of institutions from one country to another country (i.e., “professor” is “catedrático” in Spain).
\end{itemize}
\item Omission: The highlighted source part is omitted from the translation. An omission could correspond to a) a creative solution (for example, that text was omitted because it does not make sense in the translation) or b) a shortcut solution (for example, that text is omitted because it is rather cumbersome to render). 
\item Creative Shift (CS): translations that deviate from a literal rendering of the source text to create a more impactful, natural, or culturally appropriate translation in the target language in a meaningful or stylistically motivated way. 
\begin{itemize}
    \item Abstraction: Uses a more general or vague expression. E.g., Superordinate Term (e.g., “bird” for “sparrow”). Paraphrase (rephrased in general terms).
        \begin{itemize}
            \item Superordinate Term: such as hyponymy or meronym, that is a term that corresponds to a more general category, i.e., “un Rioja” translated as “a wine.”
            \item Paraphrase: the words are modified, but the underlying sense is kept, i.e., “un Rioja” translated as “a strong wine”.
        \end{itemize}
\item Concretisation: Makes the idea/image more explicit or detailed. E.g., Addition (adds extra detail); Completion (fills in implied meaning).
\begin{itemize}
\item Addition: the TT text is replaced by certain aspects of the ST, i.e., “a wine” is replaced by “un Rioja”.
\item Completion: the TT text completes the ST with information that would be implicit in the SC, i.e., “Charles Lotton” for “el actor estadounidense Charles Lotton”.
\end{itemize}
\item Modification: Uses a different image or metaphor without becoming more abstract or concrete.
\begin{itemize}
    \item Cultural: the TT uses an alternative cultural reference from the target culture, i.e., “at New York University” is replaced by “en la Universidad Autónoma de Madrid”
    \item Situational: something different in the TT that fits the situation, i.e., “in the 1960s series The Adams Family” is replaced by “la serie de los años 60 La familia Monster” for a TT in Spain.
    \item Historical: the TT is adapted to a different time, i.e., this would be the case of using inclusive language in the TT.
\end{itemize}
\end{itemize}
\item Uncertain
\end{itemize}
\noindent\textbf{Step 3: Provide reasoning}
Briefly explain why you chose that category.\\
\noindent\textbf{Step 4: Confidence score}
Assign a confidence score (1–5).
\begin{itemize}
    \item  5 = Very confident → clear and unambiguous.
    \item 3 = Unsure → some ambiguity, but decision possible.
    \item 1 = Cannot decide → too unclear to judge.
\end{itemize}

\subsubsection{Automatic evaluation}\label{sec:automatic_eval}
Tables \ref{tab:apdx_eval1}, \ref{tab:apdx_eval2}, and \ref{tab:apdx_eval3} show the prompts used in the LLM-as-a-judge setup. We evaluate by sampling 12.8\% of the human annotations, comparing translation-technique labels, and reporting macro F1 scores to account for label imbalance. Table \ref{tab:validation} shows per-label precision, recall, F1, and the confusion matrix.

\zr{At the segment level, the LLM-as-a-judge classifier shows solid overall performance, with a macro F1 of 0.710 (precision 0.722, recall 0.704) across the three labels. Performance is strongest for Reproduction (F1 0.895), while Creative Shift is identified with moderate reliability (F1 0.700). The most challenging category is Not acceptable (F1 0.533). At the system level, ranking performance is substantially stronger: the model achieves a Spearman rank correlation of 0.799 with human rankings, indicating that it is more reliable for comparing systems overall than for making fine-grained judgments on individual segments. This is consistent with the fact that the resulting difference in overall creativity score is small, at only 0.04. Taken together, these results give us reasonable confidence in using the classifier for system-level comparisons, while suggesting greater caution when interpreting segment-level creativity labels.}

\begin{table}[!htp]\centering\scriptsize
\begin{tabular}{llll}\toprule
\textbf{Class} &\textbf{Precision} &\textbf{Recall} &\textbf{F1}  \\\midrule
Reproduction &0.887 &0.904 &0.895\\
Creative Shift &0.778 &0.636 &0.700  \\
Not acceptable &0.500 &0.571 &0.533  \\
Macro &0.722 &0.704 &0.710  \\
\midrule
\multicolumn{4}{c}{\textbf{Confusion Matrix}} \\
\midrule
\textbf{Gold/Pred} &\textbf{Reproduction} &\textbf{Creative Shift} &\textbf{Not acceptable}  \\
\midrule
Reproduction &47 &1 &4  \\
Creative Shift &4 &7 &0 \\
Not acceptable &2 &1 &4  \\
Sum &53 &9 &8  \\
\bottomrule
\end{tabular}
\caption{\zr{Per-label Task 1 performance and confusion matrix on the validation set. The top panel reports precision, recall, and F1 for each label, together with macro-averaged scores. The bottom panel shows the confusion matrix, with gold labels as rows and model predictions as columns.}}\label{tab:validation}
\end{table}

\begin{table}
    \centering
    \begin{tabularx}{0.49\textwidth}{X}
    \toprule
    You are an expert in literary translation studies.\\
Your task is to classify how a specific HIGHLIGHTED segment in a literary SOURCE text
has been translated into the TARGET language. You must focus ONLY on:\\
- the highlighted source span\\
- its corresponding translated span extracted from the target translation \\
\\
Ignore the rest of the sentence except when it is strictly necessary for context.\\
--------------------------------\\
OUTPUT LABELS (YOU MUST PRODUCE ALL FOUR)\\
--------------------------------\\
1. Label: Translation Technique (choose exactly ONE)\\
   - Creative Shift (CS): translations that deviate from a literal rendering of the source text to create a more impactful, natural, or culturally appropriate translation in the target language in a meaningful or stylistically motivated way. Please make sure the deviation of meaning or image is rooted in the source. If the source meaning is not reflected or much off, this choice should not be labeled as a creative shift but unacceptable (not applicable). Please check the inner logic of the text; if the logic is wrong, then such a translation is not acceptable. \\
   - Reproduction: The translation conveys the same idea or image as the highlighted source. May still vary in style or acceptability, but the meaning is preserved. \\
   - Omission \\
   - Not acceptable (Error): The translation contains grammar errors of various types or is overly literal. E.g., literally translating "raining cats and dogs" to its mere semantic equivalent “dropping dogs and cats” rather than "heavy rain." \\
   - Not acceptable (Not Applicable): If the translation cannot be evaluated as a valid solution (e.g., distorts meaning beyond recognition but does not contain grammar issues, “too creative”; not acceptable to the target audience at all), mark it as Not Applicable (NA). E.g., Raining cats and dogs → A storm of fur filling the heavens. \\
   - Uncertain\\
    \bottomrule
    \end{tabularx}
    \caption{LLM evaluation prompt (1)}
    \label{tab:apdx_eval1}
\end{table}

\begin{table}
    \centering
    \begin{tabularx}{0.49\textwidth}{X}
    \toprule
    2. Level\_of\_creativity $\in$ [low, medium, high]: How surprising, inventive, or non-routine the translation choice feels within the given literary context. \\
- low    → routine, literal, unsurprising\\
- medium → some interpretive or stylistic choice\\
- high   → clearly inventive, or unexpected \\
    3. Level\_of\_acceptability $\in$ [low, medium, high]:  Overall translation quality and appropriateness in context: \\
     - low    → clearly unacceptable as a translation\\
     - medium → understandable but flawed, awkward, or unstable\\
     - high   → logical, fluent, appropriate, and acceptable for literary use based on the source text. Make sure the word choice is conventional for the audience, and make sure it adheres to the source meaning without added information that is not reflected or necessary in the source text. Explain the reasoning why the acceptability is high. \\
4. Level of deviation $\in$ [low, medium, high]:   \\
- low    → meaning \& imagery preserved \\
- medium → meaning \& imagery partially preserved \\
- high   → strongly reimagined meaning or image, with added information\\
!! Please double-check to ensure the deviation adheres to the source context. Be cautious in cases of medium or high deviation, as this usually indicates changes that go beyond the source meaning and are sometimes unacceptable according to the source context and label definition. Be very cautious with added meanings that are not necessary! If the deviation is labeled medium or high, explain exclusively why this choice is acceptable. \\
\\
!! If a translation is labeled as a creative shift, verify that it does not introduce excessive added meaning that cannot be inferred from the source or supported by the context. If it does, the translation is “too creative” and should be marked as unacceptable (not applicable). Likewise, if the translation is illogical or does not make sense in the source context, it is also not acceptable.
\\
    \bottomrule
    \end{tabularx}
    \caption{LLM evaluation prompt (2)}
    \label{tab:apdx_eval2}
\end{table}

\begin{table}
    \centering
    \begin{tabularx}{0.48\textwidth}{X}
    \toprule
\textbf{source}:
\$source \\
\textbf{highlighted source spans}:
\$ucp \\
--------------------------------\\
Example evaluation result: \\
--------------------------------\\
\$example \\
--------------------------------\\
Translation to be analysed\\
--------------------------------\\
\textbf{translation}:
\$translation\\
-------------------------------- \\
OUTPUT FORMAT (STRICT)\\
--------------------------------\\
Return your answer in the following JSON-like format ONLY:
{"highlighted\_source": "<highlighted span1>",
    "translated\_span": "<translated span1>",
    "label": "<Label>",
    "creativity\_level": "<Level\_of\_creativity>",
    "acceptability\_level": "<Level\_of\_acceptability>",
    "deviation\_level": "<Level\_of\_deviation>",
    "reasoning": "<reasoning>"} \\
    \bottomrule
    \end{tabularx}
    \caption{LLM evaluation prompt (3)}
    \label{tab:apdx_eval3}
\end{table}

\begin{figure*}
    \centering
    \includegraphics[width=0.99\linewidth, angle=0,origin=c, ,trim=0 0 0 0,clip]{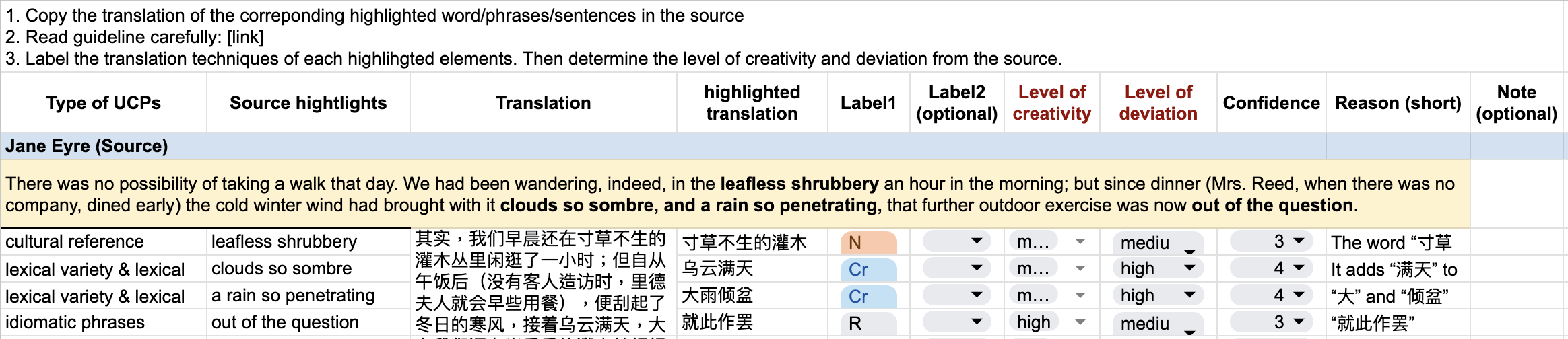}
    \caption{Screenshot of the task 2 annotation tool.}
    \label{fig:screenshot}
\end{figure*}

\subsubsection{Creative-oriented translation prompts}
\textbf{Base}: \\
Please translate the following literary excerpt into \{language\}. 

\noindent Excerpt: \{excerpt\}

\noindent Output only the \{language\} translation.

\noindent\textbf{P1}: You are a literary translator. Your goal is to translate the source text into \{language\} in a way that preserves its meaning, tone, stylistic effects, imagery, voice, and cultural nuance, while producing a fluent and natural target-language text. 

\noindent Excerpt: \{excerpt\}

\noindent Output only the \{language\} translation.

\noindent\textbf{P2}:
You are a literary translator. Your goal is to translate the source text into \{language\} in a way that preserves its meaning, tone, stylistic effects, imagery, voice, and cultural nuance, while producing a fluent and natural \{language\} text. You should balance fidelity to the original with selective creative decisions that help recreate literary effects for \{language\} readers. When a literal rendering would lose nuance or sound awkward, you may adjust expressions or imagery that suits the context.

\noindent Excerpt: \{excerpt\}

\noindent Output only the \{language\} 

\noindent\textbf{P3}: 
You are a literary translator. This source text is highly creative, and your translation should reflect that creativity in \{language\}. You may employ Creative Shifts, meaning purposeful departures from a literal rendering that create a more impactful, natural, or culturally resonant translation whenever they meaningfully enhance the literary effect. Maintain the core meaning and emotional intent of the original while allowing yourself broad stylistic freedom.

\noindent Excerpt: \{excerpt\}

\noindent Output only the \{language\} 

\subsubsection{Model configurations}\label{sec:config}
\paragraph{Temperature}
For Task~2 \zr{translation generation}, we set the decoding temperature to 0.7 to allow for some randomness, in line with recommendations for creative generation tasks~\cite{peeperkorn2024temperature}. For Task~2 auto-evaluation, we use a temperature of 0 for the selected evaluator to ensure reproducibility. For Task~1 \zr{source comprehension} on creative texts, we use a lower temperature of 0.3 to reduce randomness while still preserving the model's capacity to process literary content.

\paragraph{Model sizes} 
For our experiments, we evaluate a broad set of contemporary LLMs. From AI21, we include Jamba\mbox{-}Large and Jamba\mbox{-}mini. From Anthropic, we use Claude3.7\mbox{-}Sonnet and its reasoning variant, Claude3.7\mbox{-}Sonnet\mbox{-}Thinking. We incorporate two DeepSeek models (Deepseek\mbox{-}R1, Deepseek\mbox{-}V3.2) as well as Google’s Gemini2.5\mbox{-}Flash\mbox{-}lite and Gemini2.5\mbox{-}Flash. The Meta-LLaMa family is represented by LLaMa3.1\mbox{-}405b and LLaMa3.3\mbox{-}70b. From Mistral AI, we evaluate Mistral\mbox{-}Large, Mistral\mbox{-}Small, and Mistral\mbox{-}Small\mbox{-}Creative. We also include OpenAI models (GPT\mbox{-}4o, GPT\mbox{-}4o\mbox{-}mini, GPT\mbox{-}5.1) and a diverse set from Qwen models (Qwen2.5\mbox{-}7B, Qwen3\mbox{-}235B w/o Thinking, Qwen3\mbox{-}30B w/o Thinking). Finally, we include X\mbox{-}AI’s Grok3 and Grok3\mbox{-}mini.

\begin{figure}
    \centering
\includegraphics[width=1\linewidth]{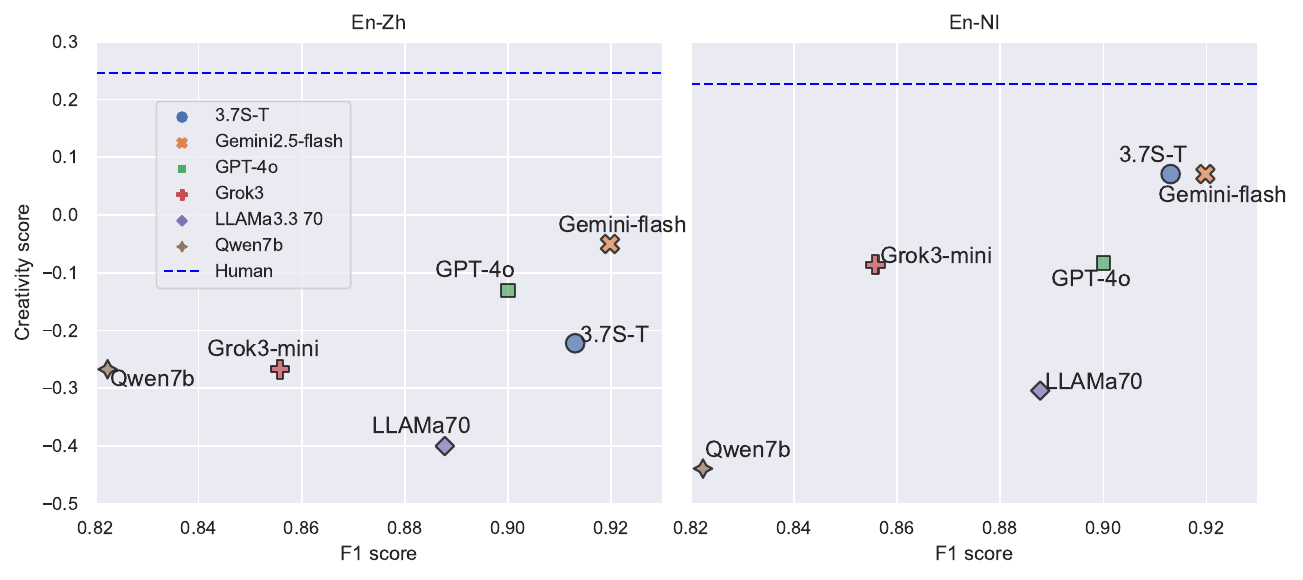}
    \caption{From source comprehension (Task~1, F1) to creative transfer (Task~2, creativity score). Creativity scores are based on human annotations of translation outputs, and the blue horizontal line denotes the creativity score of published human translations.}
    \label{fig:relation_human}
\end{figure}

\subsubsection{\zr{Qualitative examples for Task 2: Human vs. LLMs translations}}\label{sec:apdx_example}
\begin{table*}[!htp]\centering\scriptsize
\begin{tabular}{p{0.15\linewidth}p{0.15\linewidth}p{0.11\linewidth}p{0.11\linewidth}p{0.095\linewidth}p{0.215\linewidth}}
\toprule
Source &Translation &UCP &UCP Translation &Label &Reason \\\midrule
\multirow{14}{=}{1801.— I have just returned from a visit to my landlord—the solitary neighbour that I shall be troubled with. This is certainly a beautiful country! In all England, I do not believe that I could have fixed on a situation so completely removed from the stir of society. A perfect misanthropist’s heaven: and Mr. Heathcliff and I are such a suitable pair to divide the desolation between us. A capital fellow! } &\multicolumn{5}{c}{\textbf{Human}} \\
\cmidrule{2-6}
&\multirow{6}{=}{\begin{CJK*}{UTF8}{gbsn}一八〇一年。我刚刚拜访过我的房东回来——就是那个将要给我惹麻烦的孤独的邻居。这儿可真是一个美丽的乡间！在整个英格兰境内，我不相信我竟能找到这样一个能与尘世的喧嚣完全隔绝的地方，一个厌世者的理想的天堂。而希刺克厉夫和我正是分享这儿荒凉景色的如此合适的一对。一个绝妙的人！\end{CJK*}} &solitary neighbour &\begin{CJK*}{UTF8}{gbsn}孤独的邻居\end{CJK*} &Reproduction &Translating solitary into \begin{CJK*}{UTF8}{gbsn}``孤独''\end{CJK*} is ok. \\
\cmidrule{3-6}
& &the stir of society &\begin{CJK*}{UTF8}{gbsn}尘世的喧嚣\end{CJK*} &Creative Shift &Use \begin{CJK*}{UTF8}{gbsn}``尘世'' for society is a good choice in this context. Also, ``尘世'' and ``喧嚣'' forms good collocation. \end{CJK*}\\
\cmidrule{3-6}
& &A perfect misanthropist’s heaven &\begin{CJK*}{UTF8}{gbsn}一个厌世者的理想的天堂 \end{CJK*}&Reproduction &It's ok. \\
\cmidrule{3-6}
& &Mr. Heathcliff &\begin{CJK*}{UTF8}{gbsn}希刺克厉夫 \end{CJK*}&Reproduction & \begin{CJK*}{UTF8}{gbsn}I think the use of ``刺'' and ``厉'' is to echo with the imagery of heathcliff. \end{CJK*}\\
\cmidrule{3-6}
& & divide the desolation between us &\begin{CJK*}{UTF8}{gbsn}分享这儿荒凉景色\end{CJK*} &Reproduction &\begin{CJK*}{UTF8}{gbsn}It adds ``景色'' to the translation of desolation, which is creative and helps reader's understanding.\end{CJK*} \\
\cmidrule{3-6}
& &A capital fellow &\begin{CJK*}{UTF8}{gbsn}一个绝妙的人\end{CJK*} &Reproduction &\begin{CJK*}{UTF8}{gbsn}``绝佳的'' expressed the connotation of capital.\end{CJK*} \\
\cmidrule{2-6}
&\multicolumn{5}{c}{\textbf{GPT-4o}} \\
\cmidrule{2-6}
&\multirow{6}{=}{\begin{CJK*}{UTF8}{gbsn}1801年——我刚刚从拜访我的房东回来——他将是唯一会给我带来烦恼的邻居。这里确实是一个美丽的\textcolor{red}{国家}！我相信在整个英格兰，\textcolor{red}{我}都找不到如此完全远离社会喧嚣的地方。这是一个完美的厌世者的天堂：而我和希斯克利夫先生这样的搭档正好可以共同分担这份荒凉。真是个了不起的家伙！\end{CJK*}} &solitary neighbour &\begin{CJK*}{UTF8}{gbsn}唯一...的邻居 \end{CJK*}&Not acceptable (Not Applicable) & \begin{CJK*}{UTF8}{gbsn}Translating solitary into ``唯一'' (single) is not acuurate.\end{CJK*} \\
\cmidrule{3-6}
& &the stir of society &\begin{CJK*}{UTF8}{gbsn}社会喧嚣 \end{CJK*}&Reproduction & \begin{CJK*}{UTF8}{gbsn}Translating stir to ``喧嚣'' is a creative choice. The collocation ``社会喧嚣'' is not that good, but it's acceptable. \end{CJK*}\\
\cmidrule{3-6}
& &A perfect misanthropist’s heaven &\begin{CJK*}{UTF8}{gbsn}一个完美的厌世者的天堂 \end{CJK*} &Reproduction &It's ok. \\
\cmidrule{3-6}
& &Mr. Heathcliff &\begin{CJK*}{UTF8}{gbsn}希斯克利夫先生 \end{CJK*}&Reproduction &It's simply the transliteration. \\
\cmidrule{3-6}
& & divide the desolation between us &\begin{CJK*}{UTF8}{gbsn}共同分担这份荒凉 \end{CJK*}& Not acceptable (Not Applicable) & \begin{CJK*}{UTF8}{gbsn} I think this is unacceptable. ``分担'' is used for something negative, but here the desolation is not. \end{CJK*}\\
\cmidrule{3-6}
& &A capital fellow & \begin{CJK*}{UTF8}{gbsn}真是个了不起的家伙 \end{CJK*} &Reproduction & \begin{CJK*}{UTF8}{gbsn}Capital does mean excellent, but here it's not enough to just translate it into ``了不起的''. It has more contextual meaning. \end{CJK*}\\
\bottomrule
\end{tabular}
\caption{Example translation excerpt from Wuthering Heights for English--Chinese (En--Zh) Task 2, comparing human translation and GPT-4o outputs. Under the proposed creativity scoring scheme, the human translation obtains a score of 0.17 ($1/6$), while GPT-4o is assigned a score of $-0.33$ ($-2/6$). Other translation errors are highlighted in \textcolor{red}{red}.
}\label{tab:apdx_zhexamples}
\end{table*}
Table~\ref{tab:apdx_zhexamples} presents an En--Zh example based on human evaluation. It shows that, at the UCP level, the human translation relies largely on reproduction, but does so with greater precision, contextual sensitivity (e.g., by echoing the character's imagery), and occasional creative judgment, even when these choices are not formally labeled as creative shifts. Together, these qualities provide a strong foundation for creative adaptation. By contrast, GPT-4o still produces contextual interpretation errors that are unacceptable in this setting and often sound unnatural to native readers, leaving little room for genuine creative adaptation. At the full-translation level, GPT-4o also exhibits more errors overall and tends to be more literal and less stylistically consistent than the human translation.

\begin{table*}[!htp]\centering\scriptsize
\begin{tabular}{p{0.13\linewidth}p{0.21\linewidth}p{0.11\linewidth}p{0.11\linewidth}p{0.095\linewidth}p{0.17\linewidth}}
\toprule
Source &Translation &UCP &UCP Translation &Label &Reason \\\midrule
\multirow{14}{=}{There was no possibility of taking a walk that day. We had been wandering, indeed, in the leafless shrubbery an hour in the morning; but since dinner (Mrs. Reed, when there was no company, dined early) the cold winter wind had brought with it clouds so sombre, and a rain so penetrating, that further outdoor exercise was now out of the question.} &\multicolumn{5}{c}{\textbf{Human}} \\
\cmidrule{2-6}
&\multirow{4}{=}{Het was die dag niet mogelijk om te gaan wandelen. Weliswaar hadden we ’s morgens een uur door het kale struikgewas gedwaald, maar sinds het middageten (mevrouw Reed at vroeg als er geen bezoek was) had de koude winterwind zulke donkere wolken en zo’n doordringende regen meegevoerd dat er nu geen sprake meer van kon zijn om wat beweging buitenshuis te nemen.} &leafless shrubbery	& kale struikgewas	&Reproduction	&rather literal translation in Dutch; perfectly acceptable \\
\cmidrule{3-6}
&&clouds so sombre &zulke donkere wolken &Creative Shift &donker = dark instead of 'sombre'; structure itself cannot be retained in Dutch, so I did not take that into account \\
\cmidrule{3-6}
&&a rain so penetrating &zo’n doordringende regen &Reproduction &literal translation apart from structure (which cannot be retained in Dutch) \\
\cmidrule{3-6}
&&out of the question &geen sprake meer van &Reproduction &I labeled it as 'reproduction; as it is a multiword expression that is always translated as such. No idea how to label it as a UCP \\
\cmidrule{2-6}
&\multicolumn{5}{c}{\textbf{Claude3.7-Sonnet}} \\
\cmidrule{2-6}
&\multirow{4}{=}{Er was geen mogelijkheid om die dag een wandeling te maken. We hadden inderdaad 's ochtends een uur in de kale struiken rondgedwaald; maar sinds het middageten (mevrouw Reed dineerde vroeg wanneer er geen gezelschap was) had de koude winterwind zulke sombere wolken en zo'n doordringende regen meegebracht, dat verdere buitenactiviteit nu uitgesloten was.} &leafless shrubbery &kale struiken &Reproduction & ``struiken'' = ``bushes''; is almost same image; deviates a little bit \\
\cmidrule{3-6}
&&clouds so sombre &zulke sombere wolken &Reproduction &literal translation apart from structure (which cannot be retained in Dutch) \\
\cmidrule{3-6}
&& a rain so penetrating &zo'n doordringende regen &Reproduction &literal translation apart from structure (which cannot be retained in Dutch) \\\cmidrule{3-6}
& &
out of the question &uitgesloten &Creative Shift &uitgesloten = impossible \\
\bottomrule
\end{tabular}
\caption{Example translation excerpt from Jane Eyre for English--Dutch (En--Nl) Task 2, comparing human translation and Claude3.7-Sonnet outputs. Under the proposed creativity scoring scheme, both translations receive a score of 0.25 ($1/4$).
}\label{tab:apdx_nlexamples}
\end{table*}
\begin{table*}[!htp]\centering\tiny
\begin{tabular}{p{0.12\linewidth}p{0.25\linewidth}p{0.069\linewidth}p{0.105\linewidth}p{0.09\linewidth}p{0.2\linewidth}}
\toprule
Source &Translation &UCP &UCP Translation &Label &Reason \\\midrule
\multirow{14}{=}{[...] all I know is, that there was but one solitary bidding, and that was from an attorney connected with the bill-broking business, who offered two pounds in cash, and the balance in sherry, but declined to be guaranteed from drowning on any higher bargain. Consequently the advertisement was withdrawn at a dead loss--for as to sherry, my poor dear mother’s own sherry was in the market then--and ten years afterwards, the caul was put up in a raffle down in our part of the country, to fifty members at half-a-crown a head, the winner to spend five shillings.} &\multicolumn{5}{c}{\textbf{Human}} \\
\cmidrule{2-6}
&\multirow{4}{=}{\tiny{[...] Ik weet niet of de zeevarenden op dat tijdstip slecht bij kas waren of de voorkeur gaven aan zwemgordels, maar het is een feit dat er slechts één bod werd gedaan en dit wel door een wisselmakelaar, die aanbood twee pond in contanten en de rest in sherry te betalen, 
maar geen penny meer wilde geven om tegen verdrinken gewaarborgd te zijn. Dientengevolge waren de onkosten van de advertentie noodeloos gemaakt—mijne arme moeder was
in die dagen genoodzaakt haar eigen wijn zelfs te verkoopen—en tien jaar later werd de helm ergens in onze buurt onder vijftig deelnemers verloot tegen een halve kroon per hoofd, terwijl de winner nog vijf shilling zou toegeven.}} & olitary bidding &bod &Omission &solitary is not translated, it is seen as part of "één", which is correct meaningwise but it leaves out something \\
\cmidrule{3-6}
& &
bill-broking business &/ &Creative Shift &here the "attorney connected with" is translated as "wisselmakelaar" which in turn results in a change in translation for the rest of this NP, it is more abstract \\
\cmidrule{3-6}
& &
at a dead loss &noodeloos &Reproduction &I think there is a slight shift in meaning but it makes sense in Dutch so I am not too bothered by it \\
\cmidrule{3-6}
& &at half-a-crown a head &tegen een halve kroon per hoofd &Reproduction &pretty literal translation but it makes sense not to change the monetary value, "head" sounds quite old, there are better options. \\
\cmidrule{2-6}
&\multicolumn{5}{c}{\textbf{Claude3.7-Sonnet}} \\
\cmidrule{2-6}
&\multirow{4}{=}{[...] het enige wat ik weet is dat er slechts één enkel bod kwam, en dat was van een advocaat verbonden aan het wisselmakelaarsbedrijf, \textcolor{blue}{die twee pond contant bood en de rest in sherry, 
maar weigerde tegen een hogere prijs gegarandeerd te worden tegen verdrinking. Bijgevolg werd de advertentie ingetrokken met een volledig verlies - want wat betreft sherry,} de eigen sherry van mijn arme lieve moeder stond toen al te koop - en tien jaar later werd de helm verloot in ons deel van het land, aan vijftig deelnemers tegen een halve kroon per persoon, \textcolor{blue}{waarbij de winnaar vijf shilling moest uitgeven}. }&solitary bidding &enkel bod &Reproduction &great translation \\
\cmidrule{3-6}
& &
bill-broking business &wisselmakelaarsbedrijf &Reproduction &pretty literal translation, again I don't know why "wisselhandel" is not used since that is the most used term \\
\cmidrule{3-6}
& &at a dead loss &met een volledig verlies &Reproduction &good translation, I like the alliteration \\
\cmidrule{3-6}
& &at half-a-crown a head &tegen een halve kroon per persoon &Reproduction &pretty literal translation but it makes sense not to change the monetary value. I also like the use of "persoon" instead of "hoofd", it sounds much better \\
\cmidrule{2-6}
&\multicolumn{5}{c}{\textbf{Gemini2.5-Flash}} \\
\cmidrule{2-6}
&\multirow{4}{=}{[...] alles wat ik weet is dat er maar één enkel bod was, en dat kwam van een advocaat die verbonden was aan de wisselmakelaardij, 
\textcolor{blue}{die twee pond contant bod en het restant in sherry, 
maar weigerde om bij een hogere deal gegarandeerd te worden tegen verdrinking.
Bijgevolg werd de advertentie ingetrokken met een totaal verlies—want wat sherry betreft, }de eigen sherry van mijn arme dierbare moeder was toen al op de markt—en tien jaar later werd de helm in onze streek verloot onder vijftig leden voor tweeënhalve shilling per persoon, \textcolor{blue}{waarbij de winnaar vijf shilling moest besteden.}} & solitary bidding &enkel bod &Reproduction &great translation \\
\cmidrule{3-6}
& &
bill-broking business &wisselmakelaardij &Not acceptable (Error) &wisselmakelaardij doesn't exist, I think it's meant to be "wisselmakelarij" but I don't know why "wisselhandel" is not used since that is the most used term \\
\cmidrule{3-6}
& &
at a dead loss &met een totaal verlies &Reproduction &good expression in Dutch \\
\cmidrule{3-6}
& &at half-a-crown a head &voor tweeënhalve shilling per persoon &Creative Shift &the translation changes "half-a-crown" to another British monetary value "shilling" that is maybe slightly better known by Dutch speaker, so it makes sense. \\
\bottomrule
\end{tabular}
\caption{Example translation excerpt from David Copperfield for English--Dutch (En--Nl) Task 2, comparing human translation and the outputs of Claude3.7-Sonnet and Gemini2.5-Flash. Under the proposed creativity scoring scheme, the human translation receives a score of 0.25 ($1/4$), whereas both model outputs receive a score of 0. Translations with high similarity are highlighted in \textcolor{blue}{blue}.
}\label{tab:apdx_nlexamples2}
\end{table*}
Table~\ref{tab:apdx_nlexamples} and Table~\ref{tab:apdx_nlexamples2} present En--Nl translation examples evaluated by human annotators. In the previous example, both the human translation and Claude3.7-Sonnet receive the same creativity score, with outputs largely relying on reproduction at the UCP level.

In contrast, in Table~\ref{tab:apdx_nlexamples2}, the human translation outperforms both model outputs. A key difference lies in the greater syntactic and lexical diversity exhibited by the human translator. The human version demonstrates more flexible restructuring, varied phrasing, and contextually appropriate lexical choices. By comparison, both Claude3.7-Sonnet and Gemini2.5-Flash produce translations that are more homogenized, adhering closely to the source structure and favoring more literal or conventional formulations. This results in less variation at both the syntactic and lexical levels, and fewer instances of creative potential.

Overall, all examples highlight that while model outputs can match human performance in low-creativity settings, they still struggle to achieve the same level of diversity, adaptability, and creativity.

\begin{table*}[!htp]\centering\small
\begin{tabular}{lrrr}\toprule
\multirow{2}{*}{Model} &\multicolumn{2}{c}{Creativity score} \\\cmidrule{2-3}
&Avg. &Max \\\midrule
Claude3.7-Sonnet &\cellcolor[HTML]{fcf2f1}-0.110 &\cellcolor[HTML]{fefcfc}-0.071 \\
Claude3.7-Sonnetonnet-thinking &\cellcolor[HTML]{fefefe}-0.061 &\cellcolor[HTML]{caeada}0.012 \\
Deepseek-R1 &\cellcolor[HTML]{b6e2cc}0.039 &\cellcolor[HTML]{9ed8bc}0.071 \\
Deepseek-V3.2-exp &\cellcolor[HTML]{c6e8d7}0.018 &\cellcolor[HTML]{a7dcc2}0.060 \\
Gemini2.5-Flash (lite) &\cellcolor[HTML]{fbebe9}-0.140 &\cellcolor[HTML]{fdf6f5}-0.095 \\
Gemini2.5-Flash &\cellcolor[HTML]{f2faf6}-0.042 &\cellcolor[HTML]{c1e6d4}0.024 \\
GPT4o &\cellcolor[HTML]{fefefd}-0.064 &\cellcolor[HTML]{ddf1e7}-0.014 \\
GPT4o-mini &\cellcolor[HTML]{f8ddda}-0.196 &\cellcolor[HTML]{f9e4e2}-0.167 \\
GPT5.1 &\cellcolor[HTML]{fdf7f7}-0.090 &\cellcolor[HTML]{fdf9f8}-0.083 \\
Grok3 &\cellcolor[HTML]{fbfefc}-0.054 &\cellcolor[HTML]{caeada}0.012 \\
Grok3-mini &\cellcolor[HTML]{fcf4f3}-0.104 &\cellcolor[HTML]{dcf1e6}-0.012 \\
Jamba1.7-Large &\cellcolor[HTML]{f1b7b2}-0.348 &\cellcolor[HTML]{f4c9c6}-0.274 \\
Jamba1.7-Mini &\cellcolor[HTML]{e67c73}-0.586 &\cellcolor[HTML]{e7857d}-0.548 \\
Llama3.1-405b &\cellcolor[HTML]{f7dad7}-0.208 &\cellcolor[HTML]{f9e4e2}-0.167 \\
Llama3.3-70b &\cellcolor[HTML]{f6d2cf}-0.238 &\cellcolor[HTML]{f9e1df}-0.179 \\
Mistral-Large &\cellcolor[HTML]{95d4b5}0.083 &\cellcolor[HTML]{57bb8a}0.167 \\
Mistral-Small &\cellcolor[HTML]{fbedec}-0.130 &-0.060 \\
Mistral-Small-Creative &\cellcolor[HTML]{dcf1e6}-0.012 &\cellcolor[HTML]{84cda9}0.107 \\
Qwen2.5-7b &\cellcolor[HTML]{f6d1ce}-0.243 &\cellcolor[HTML]{f9e4e2}-0.167 \\
Qwen3-235b &\cellcolor[HTML]{fefdfd}-0.065 &\cellcolor[HTML]{edf8f3}-0.036 \\
Qwen3-235b-Thinking &\cellcolor[HTML]{afdfc8}0.048 &\cellcolor[HTML]{83cda9}0.108 \\
Qwen3-30b &\cellcolor[HTML]{b2e0c9}0.045 &\cellcolor[HTML]{a7dcc2}0.060 \\
Qwen3-30b-Thinking &\cellcolor[HTML]{ebf7f1}-0.033 &\cellcolor[HTML]{c1e6d4}0.024 \\
\bottomrule
\end{tabular}
\caption{Creativity scores of models across four prompts. “Avg.” denotes performance averaged over the four prompts, while “Max” refers to the best performance achieved among the prompts.}\label{tab:scores}
\end{table*}
\subsubsection{Creativity scores across four prompts}
Table~\ref{tab:scores} contains the creativity scores of the models per automatic annotation across four prompts. “Avg.” denotes performance averaged over the four prompts, while “Max” refers to the best performance achieved among them.

\subsection{\zr{Adversarial study: memorization effects in Task 1}}\label{apdx:ablation}
We run an adversarial evaluation to test whether Task 1 claim judgments and reasonings are grounded in the provided paragraph rather than in memorized source knowledge. We sample 25 claims from Jane Eyre, Frankenstein, Mrs Dalloway, and Animal Farm and evaluate 8 models, yielding 200 cases in total. For each case, we adversarially modify the paragraph and remove related background context, so that the model must rely only on the modified paragraph and the question.

These edits preserve surface plausibility while breaking alignment with the original text. We use two attack types:

Entity substitution. We replace salient details such as names, locations, and book-specific references; for example, Jane to Mary, Geneva to Germany, or Bewick’s History of British Birds to Wick’s History of British Elephants. The paragraph remains fluent, but key factual anchors are altered.
Plot or attribute swap. We modify key character or event properties, such as reassigning actions, exchanging appearances, or changing a character’s gender. In Animal Farm, for example, this can reassign traits or roles in a locally coherent way that conflicts with the original story.

We evaluate GPT-4o, GPT-5.1, Claude3.7-Sonnet, Claude3.7-Sonnet-thinking, Grok3, DeepSeek-R1, Mistral-Small-Creative, and Mistral-Large. We measure (1) whether the Task 1 label changes under adversarial perturbation and (2) whether the explanation includes unsupported references to the original plot or characters, which we treat as evidence of memorization.

Label changes are rare. Out of 200 cases, we observe 2 label changes for Claude3.7-Sonnet, DeepSeek-R1, and GPT-4o; 1 for Claude3.7-Sonnet-thinking, GPT-5.1, Grok3, and Mistral-Large; and 0 for Mistral-Small-Creative, corresponding to an overall rate of 5.0\% (10/200). Memorization-like effects are similarly infrequent: unsupported references occur in 1 case for Claude3.7-Sonnet and Grok3, and in 2 cases for DeepSeek-R1, GPT-4o, Mistral-Large, and Mistral-Small-Creative; this again corresponds to 10/200 cases (5.0\%). Table \ref{tab:apdx_ablation} presents three representative examples illustrating both evidence-based and memorization-affected reasoning under the same judgment label.

\begin{table*}[!htp]\centering\small
\begin{tabularx}{0.985\textwidth}{llX}
\toprule
\textbf{Model} & \textbf{Mode} & \textbf{Reasoning} \\\midrule
\multirow{2}{*}{GPT-5.1} & adversarial &The quote states that \textcolor{green}{Mary cares a lot for} the “letterpress” of \textcolor{green}{Wick’s History of British Elephants} but sometimes skips certain introductory pages. “Letterpress” historically refers to the printed text (as opposed to illustrations or plates), not to a particular reading habit that excludes images. Nothing in the passage says she only reads the letterpress and ignores illustrations; illustrations are not even mentioned. The claim adds a contrast (text vs. images) that is not grounded in the quote. \\
\cline{2-3}
&original &\textit{Jane} explicitly chooses “a volume, taking care that it should be one stored with pictures,” and then says “the letterpress thereof \textit{I cared little for}.” She prefers the illustrations and cares little for the text, the opposite of only reading letterpress and ignoring images. \\
\midrule
\multirow{2}{*}{Claude3.7-Sonnet} & adversarial &This claim contains a \textcolor{green}{factual error} - the white stripe is on \textcolor{green}{Glove's mouth}, not \textcolor{green}{Boxing's}. Additionally, this physical detail is explicitly connected to \textcolor{green}{Glove's appearance} and perceived intelligence in the text: 'A white stripe down \textcolor{green}{his mouth} gave him a somewhat stupid appearance.' The physical trait directly relates to character perception and is immediately followed by commentary on his actual intelligence, showing it functions beyond mere physical description to establish character. \\
\cline{2-3}
&original &\textit{Boxer's 'white stripe down his nose}' is not merely a physical detail but directly tied to character development, as the text explicitly connects this feature to his appearance of stupidity: 'A \textit{white stripe down his nose} gave him a somewhat stupid appearance, and in fact he was not of first-rate intelligence.' This physical trait serves as visual shorthand for \textit{Boxer's intellectual limitations}, which will become significant to the plot as the story progresses. The detail helps establish \textit{Boxer}'s character as physically strong but intellectually limited, a crucial aspect of his role in the allegory. \\
\midrule
\multirow{2}{*}{Mistral-Large} & adversarial &The quote seamlessly blends external social reality (e.g., \textcolor{green}{Lucia}'s workload, \textcolor{green}{Mayer}'s men arriving, the practical task of buying \textcolor{green}{tofu}) with \textcolor{red}{Clarissa Dalloway}'s interior perception (e.g., her reflective thoughts about the day feeling 'fresh as if issued to children on a beach'). This juxtaposition of the mundane and the poetic, the external and the internal, is characteristic of \textcolor{red}{Woolf}'s stream-of-consciousness style, where disparate elements coexist in a single narrative breath.\\
\cline{2-3}
&original &The passage seamlessly blends social reality (e.g., \textit{Lucy}'s work, the removal of doors) with \textit{Clarissa}'s interior perception (e.g., her sensory and emotional response to the morning). This fusion is a defining feature of \textit{Woolf's modernist} approach. \\
\bottomrule
\end{tabularx}
\caption{Examples of model responses on original and adversarially modified inputs. \textcolor{green}{Green} highlights denote evidence-grounded reasoning, while \textcolor{red}{red} highlights denote failures, including cases where the model relies on details from the original text rather than the modified evidence. Text in \textit{italics} highlights evidence drawn from the original passage.}\label{tab:apdx_ablation}
\end{table*}

Overall, the adversarial results suggest that Task 1 is a largely valid test of context-grounded reasoning. In most cases, models derive their judgments from the provided paragraph, and adversarial perturbations rarely change the final label. At the same time, we identify a small number of cases in which explanations rely on source-text knowledge rather than the modified evidence, indicating that memorization can still affect reasoning. Taken together, these findings suggest that Task 1 primarily measures contextual comprehension, while also motivating explanation-level analysis to detect occasional effects of prior exposure.

\end{document}